\pdfoutput=1

\documentclass[11pt]{article}

\usepackage[final]{acl}

\usepackage{times}
\usepackage{latexsym}

\usepackage[T1]{fontenc}

\usepackage[utf8]{inputenc}

\usepackage{microtype}

\usepackage{inconsolata}

\usepackage{amsmath,amsfonts,bm}

\def\sref#1{\S~\ref{#1}}

\def\eqref#1{equation~\ref{#1}}

\def\1{\bm{1}}

\DeclareMathAlphabet{\mathsfit}{\encodingdefault}{\sfdefault}{m}{sl}
\SetMathAlphabet{\mathsfit}{bold}{\encodingdefault}{\sfdefault}{bx}{n}

\usepackage{framed}
\usepackage{hyperref}
\usepackage{url}
\usepackage[utf8]{inputenc} 
\usepackage[T1]{fontenc}    
\usepackage{hyperref}       
\usepackage{url}            
\usepackage{booktabs}       
\usepackage{amsfonts}       
\usepackage{nicefrac}       
\usepackage{microtype}      
\usepackage{xcolor}         
\usepackage{amsmath,amssymb,amsfonts,amsthm,mathrsfs}
\usepackage[page,title,titletoc,header]{appendix}
\usepackage{enumerate}
\usepackage{color}
\usepackage{colortbl}
\usepackage{graphicx}
\usepackage{indentfirst}
\usepackage[mathscr]{eucal}
\usepackage{multirow,makecell}
\usepackage[ruled,linesnumbered]{algorithm2e}
\usepackage{algpseudocode}

\usepackage{nicefrac}
\usepackage{subfigure}
\usepackage{enumitem}
\usepackage{graphicx}
\usepackage{wrapfig}

\usepackage{CJKutf8}
\usepackage{tabularray}
\usepackage{bigstrut}
\usepackage{bbm}
\usepackage{wrapfig}
\usepackage{float}
\usepackage{caption}
\usepackage{fontawesome}
\usepackage[capitalize]{cleveref}
\usepackage{tcolorbox}
\usepackage{longtable}
\usepackage{listings}
\usepackage{multicol}

\title{\method{JumpCoder}: Go Beyond Autoregressive Coder via Online Modification}

\author{%
  Mouxiang Chen, Hao Tian,  Zhongxin Liu\thanks{Corresponding author.}, Xiaoxue Ren, Jianling Sun\\
  The State Key Laboratory of Blockchain and Data Security,\\
  Zhejiang University \\
  \texttt{\{chenmx,icouldbe,liu\_zx,xxren,sunjl\}@zju.edu.cn}%
}

\newcommand{\method}[1]{\textsc{#1}}
\newcommand{\model}{\method{JumpCoder}{}}
\newcommand{\eg}{{\it e.g.}}

\newcommand{\ie}{{\it i.e.}}

\begin{document}

\maketitle

\begin{abstract}

While existing code large language models (code LLMs) exhibit impressive capabilities in code generation, their autoregressive sequential generation inherently lacks reversibility. This limitation hinders them from timely correcting previous missing statements during coding as humans do, often leading to error propagation and suboptimal performance. We introduce \model, a novel model-agnostic framework that enables human-like online modification and non-sequential generation to augment code LLMs. The key idea behind \model~is to insert new code into the currently generated code when necessary during generation, which is achieved through an auxiliary infilling model that works in tandem with the code LLM. Since identifying the best infill position beforehand is intractable, we adopt an \textit{infill-first, judge-later} strategy, which experiments with filling at the $k$ most critical positions following the generation of each line, and uses an Abstract Syntax Tree (AST) parser alongside the Generation Model Scoring to effectively judge the validity of each potential infill. Extensive experiments using six state-of-the-art code LLMs across multiple and multilingual benchmarks consistently indicate significant improvements over all baselines. Our code is public at \faGithub~\url{https://github.com/Keytoyze/JumpCoder}.

\end{abstract}
\begin{figure*}[ht]
  \centering
  \includegraphics[width=\textwidth]{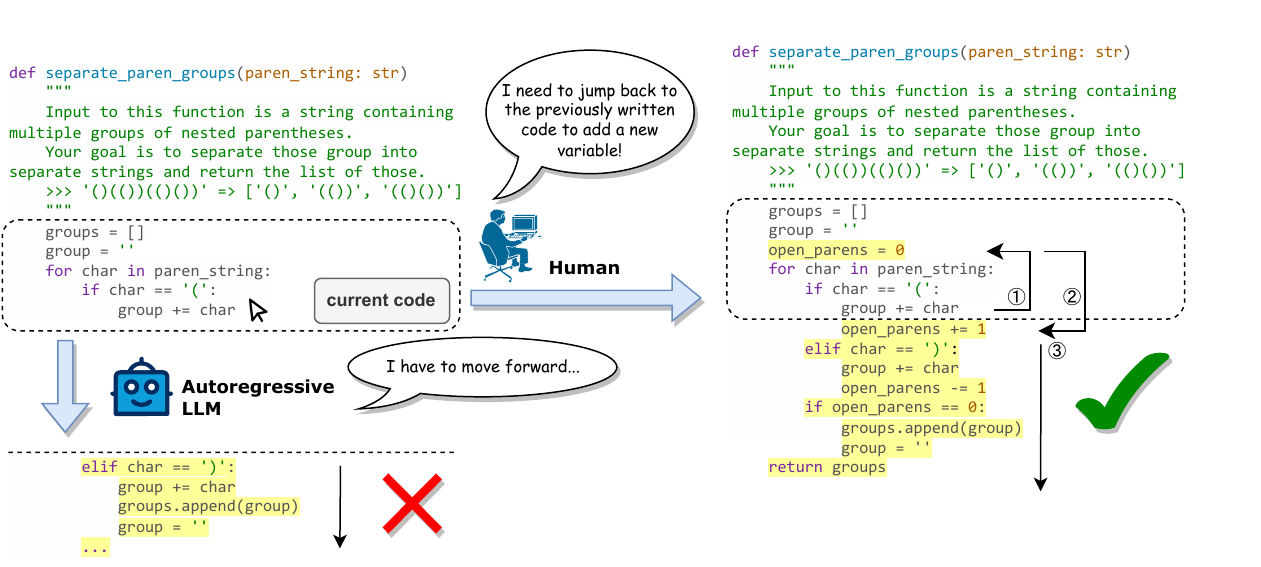}
  \caption{An illustrative example demonstrating the difference between humans and LLMs. When a new variable is required, humans can jump back to the front section to define it, but LLMs, constrained by their autoregressive nature, can only continue generation and lead to error propagation.}
  \label{fig:motivation_example}
\end{figure*}

\section{Introduction}

Recently, Large Language Models (LLMs) \cite{zhao2023survey} have attracted considerable interest and achieved notable successes. As a prominent application of LLMs, numerous code LLMs \citep{DBLP:conf/iclr/FriedAL0WSZYZL23,li_starcoder_2023,luo_wizardcoder_2023,roziere_code_2023, wang2023codet5+,wei_magicoder_2023, zheng2023codegeex} are frequently released and exhibit exceptional efficacy in code generation tasks. 
Despite their remarkable success, current LLMs work in an autoregressive style, generating text segments sequentially from left to right. This leads to an inherent limitation of \textbf{\textit{irreversibility}} --- they are incapable of revising previously generated text. 
Notably, even skilled human programmers often struggle to write code in a linear style, since real-world code commonly undergoes iterative editing and refinement \cite{DBLP:conf/iclr/FriedAL0WSZYZL23}. For example, to accommodate the developing code, programmers need to retrospectively introduce key declarations, such as new variables, references, and functions in the previously written code.

During the early stage of generation, code LLMs often miss key declarations due to the high generation uncertainty~\cite{zheng_self-infilling_2023}. Because of the irreversibility limitation, code LLMs, unlike humans, cannot add missing declarations when they need these declarations later. This often results in producing code with undefined identifiers, known as \texttt{NameError} in Python.
Notably, for the widely-used \textsc{CodeLlama} \cite{roziere_code_2023}, \texttt{NameErrors} accounted for up to 6.0\% to 18.6\% of the total error cases on HumanEval in our experiments. Although these \texttt{NameErrors} are relatively easy to address by syntax checking or model constraints \cite{dong2023codep}, the lack of necessary declarations or statements in the generated code segments often hinders LLMs from continuing to generate semantically correct code, causes the error to accumulate continuously along the generation, and limits the performance of LLMs. \cref{fig:motivation_example} presents an intuitive example, demonstrating that in the absence of necessary variables, LLMs are prone to generating syntactically correct but semantically incorrect code, which is more difficult to detect and rectify. 


\begin{figure}
    \includegraphics[width=0.45\textwidth]{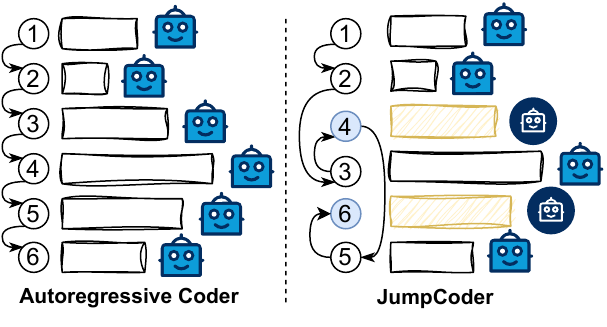}

    \caption{Schematic illustrations of traditional autoregressive coder and the proposed \model. Code lines are generated by generation model ({\raisebox{-2.5mm}{\includegraphics[width=6mm]{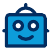}}}) and infilling model ({\raisebox{-2.5mm}{\includegraphics[width=6mm]{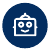}}}).}
    \label{fig:method-outline}
    \vspace{-0.6cm}
\end{figure}


To address this limitation, the crux is to enable the model to \textit{jump back} to previously generated code for inserting new code like a human whenever a missing declaration or statement is required. Based on it, we propose \textbf{\model}, a model-agnostic framework to enhance code generation performance. We introduce an innovative hybrid code generation scheme combining two models: the \textit{generation model} drafts the subsequent line of code, while the \textit{infilling model} \cite{DBLP:conf/iclr/FriedAL0WSZYZL23} fills a line within the generated code when necessary. The infilling model can retrospectively declare new variables, add new functions, insert additional calculations, and so on, enabling \model~to achieve online modification and non-sequential generation. Both the generation model and the infilling model are pre-trained and do not require further fine-tuning. \cref{fig:method-outline} illustrates the comparison between traditional autoregressive coder and our proposed \model. 

A major challenge to instantiate this idea is deciding whether (and where) to infill, or continue generation from the current code. To overcome this, we propose an \textit{infill-first, judge-later} paradigm: after generating a line, let the infilling model experiment with filling at the start of the $k$ most critical lines that have been generated, and subsequently judge their contributions to the current generation. This paradigm benefits from \textit{parallel generation} and \textit{speculative infilling} optimizations, 
which avoids a significant reduction in efficiency. 
The next challenge is how to implement the judgment method. We combine an \textbf{Abstract Syntax Tree (AST) parser} and the \textbf{Generation Model Scoring} to achieve this. AST parser is employed for the code using undefined identifiers, which accepts the infill that correctly adds the missing declaration. For other scenarios, the generation model scores the code following each infill position by comparing their mean token logit improvements. If an infill enhances the score of subsequent lines, we assume it indicates an improvement in overall generation quality. In cases outside these two scenarios, we infer that infilling is unnecessary, and opt to generate the next line. 

Extensive experiments on six variants of \method{CodeLlama} \cite{roziere_code_2023} and \method{WizardCoder} \cite{luo_wizardcoder_2023} on five code generation benchmarks indicated a consistent and significant enhancement across all baselines. Notably, \model~is adaptable to various programming languages, and helps the code LLMs achieve a pass rate increase of 4.8\% - 8.2\% in four multilingual HumanEval benchmarks at most, while substantially reducing undefined identifier errors.

To the best of our knowledge, our approach enables online modification for code generation for the first time. We hope our work can inspire further research into the irreversibility limitation of LLMs in the future. Our contributions are threefold: 

\begin{itemize}[leftmargin=*]
    \item We investigate the irreversibility, a significant yet underexplored limitation, of code LLMs.
    \item We introduce \model, a model-agnostic code generation framework for augmenting code LLMs without retraining.
    \item We perform an extensive evaluation of \model~on various program languages and code LLMs, demonstrating widespread improvements over current baseline models.
\end{itemize}
\section{Preliminaries}
\paragraph{Code generation}
Code generation, an important task of software engineering, focuses on automatically producing source code from software requirements \cite{CodeGeneration_fromRequirements}. 
The emergence of LLMs has propelled this field, demonstrating remarkable capabilities in code generation \citep{roziere_code_2023,wei_magicoder_2023,luo_wizardcoder_2023}, especially when fine-tuned on domain-specific code datasets.

\paragraph{Code infilling}
Conventional language models typically train on left-to-right next token prediction. Although LLMs excel in code generation, they face limitations in code editing tasks like code infilling, where handling bidirectional context is crucial. To overcome this, the \textbf{Fill-In-the-Middle} (FIM) training method \cite{bavarian_efficient_2022} has been proposed. FIM divides input code into three parts (\texttt{prefix}, \texttt{infix}, \texttt{suffix}), then reorders them into a sequence $\texttt{<PRE>}\oplus \texttt{{prefix}}\oplus \texttt{<SUF>}\oplus \texttt{{suffix}}\oplus \texttt{<MID>}\oplus \texttt{{infix}}\oplus \texttt{<EOT>}$ for training. This trains the model on the conditional distribution $P(\texttt{{infix}}\mid \texttt{<PRE>}\oplus \texttt{{prefix}}\oplus \texttt{<SUF>}\oplus \texttt{{suffix}})$, enhancing its infilling capabilities. However, this approach doesn't directly translate to code generation, where subsequent contexts are absent.
\section{JumpCoder}

This section elaborates on \model, which is designed to address the irreversibility limitation of autoregressive generation. Such limitation often leads to the generation of syntactically or semantically incorrect code. To address it, the key is to facilitate the model’s ability to achieve \textit{online modification}, \eg, jump back to the previously generated code to insert a new line like a human, which enables code LLMs to add missing statements. Based on it, we introduce an innovative hybrid generation approach that enables the retrospective insertion of new lines by assisting the generation model with an infilling model, as shown in \cref{fig:method-outline}. 

However, identifying the optimal locations for infilling within the current code is a significant challenge. We discovered that while pinpointing the best infill location beforehand is intractable, assessing the suitability of an infill post-insertion is comparatively simpler. Therefore, we propose an \textit{infill-first, judge-later} strategy: after generating a line, the infilling model experiments with inserting a line before each of the $k$ most critical lines that have been generated, followed by judging their impact on the current generation. This judging process is crucial as we found that unnecessary infills tend to degrade code quality (see Appendix \ref{sec:app_example} for concise examples).

\begin{figure*}[t]
  \centering
  \includegraphics[width=0.85\textwidth]{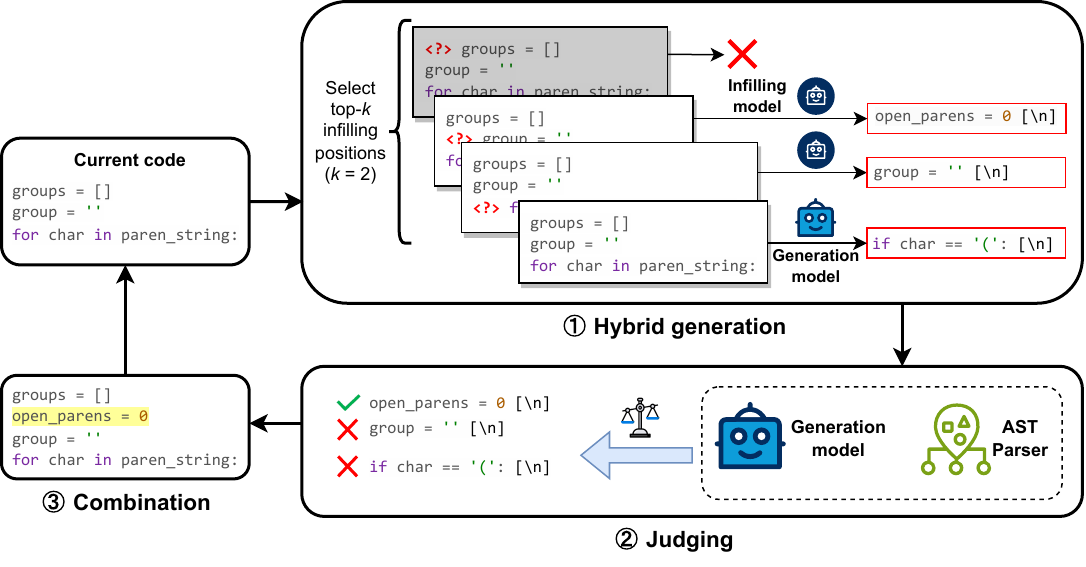}
  \caption{\model~Framework. The iterative code update process comprises three important stages: Hybrid generation, Judging and Combination. Each iteration inserts a new line of code.}
  \label{fig:method}
\end{figure*}

Specifically, \model~iteratively updates code line by line. Each iteration adds a new line to the current code, which encompasses three steps: \textbf{Hybrid generation} (\sref{sec:method-hg}), involving simultaneous line generation by both models; \textbf{Judging} (\sref{sec:method-judge}), assessing the infilling model’s output to determine line selection; and \textbf{Combination} (\sref{sec:method-combine}), merging this selection into the current code before continuing proceeding. The complete framework is outlined in \cref{fig:method}, and the algorithm is summarized in \cref{sec:app_algorithm}.

\subsection{Hybrid Generation}\label{sec:method-hg}

In this section, we describe the overview of hybrid generation in \sref{sec:method-hg-overview}, and introduce optimization techniques for speeding in \sref{sec:method-hg-optim}.

\subsubsection{Overview}\label{sec:method-hg-overview}

We consider the $n$-th iteration when we have $n$ lines of generated code. Let $(L_1, L_2, \cdots, L_n)$ denote current code, where $L_i=(x_1^{i}, x_2^{i}, \cdots, x_{|L_i|}^{i})$ denotes the $i$-th line and $x_j^{i}$ is its $j$-th token. We generate a new line $\mathbb{L}_{n+1}'$ using a generation model $\mathcal M_G$ and infill a line $\mathbb{L}_i'$ before each $L_i$ using an infilling model $\mathcal M_I$. It is important to note that a higher value of $n$ significantly increases computational and memory demands for infilling. Consequently, we limit our focus to the top-$k$ positions of \textit{uncertainty} for infilling. This uncertainty is quantified by the logit of the first non-indent token in $L_i$ given by the generation model $\mathcal M_{G}$. The underlying principle is that a lower logit for the initial non-indent token suggests a higher potential for varied generations at that position, thereby making it a prime candidate for infilling a new line. We exclude the initial indentation tokens because they are largely predetermined by syntactic rules or coding standards, which lead to elevated logit values and do not effectively reflect uncertainty for these lines. The overall process is formulated as:
\begin{align*}
    \mathcal I &= \underset{i\in \{1,\cdots,n\}}{\arg \text{Top-}k}\ S(x_{\mathbf{0}}^{i}; \mathcal M_G),\\
    \mathbb{L}_i' &\sim P_{\mathcal M_I}(\cdot \mid \texttt{<PRE>}\oplus L_{[1:i-1]}\oplus \texttt{<SUF>}\\
    &\qquad\qquad\;\;\oplus L_{[i:n]}),\quad i\in \mathcal I,\\
    \mathbb{L}_{n+1}' &\sim P_{\mathcal M_G}(\cdot \mid L_{[1:n]}),
\end{align*}
where $x_{\mathbf{0}}^{i}$ denotes the first non-indent token in line $L_i$, and $S(\cdot; \mathcal M_G)$ denotes the logit determined by $\mathcal M_G$. $P_{\mathcal M_I}(\cdot)$ and $P_{\mathcal M_G}(\cdot)$ signify sampling a line through infilling and generation respectively. 

\subsubsection{Efficiency Optimization} \label{sec:method-hg-optim}

Compared to the autoregressive method, our approach introduces an additional infilling model and extra computations, which may potentially increase overhead. Fortunately, these computations can be substantially optimized in practice, as follows:

\paragraph{Parallel Generation} Hybrid generation can do in parallel, reducing time complexity from $O(N(k+1))$ to $O(N)$ when parallelization is available.

\paragraph{Speculative Infilling}  Despite parallel processing, each line's processing time still depends on the slowest of the $k+1$ lines, typically the infill line. We found that \model~frequently infills the same positions repeatedly on different iterations, often leading to identical infilling results. Based on it, we cache the infill for each position after the generation. When infilling the same position again, we employ the speculative decoding \citep{chen2023accelerating,leviathan2023fast} to parallelly verify the cached infill's correctness and resort to standard infilling only if necessary. This strategy improves speed by around 30\%.

Despite these optimizations, one limitation is that the hybrid generation imposes extra arithmetic operations and increases memory consumption, particularly when the infilling and generation models differ. Fortunately, the bottleneck in practice often lies in memory bandwidth and communication, not arithmetic operations \cite{leviathan2023fast}, enabling the use of extra resources for hybrid generation.

\subsection{Judging}\label{sec:method-judge}

Upon acquiring $k$ infilling lines and a generation line, the next step is to judge which one from these $k+1$ lines is the most appropriate for updating the existing $n$ lines. This judging process is accomplished by integrating an \textbf{Abstract Syntax Tree (AST) Parser} with the \textbf{Generation Model Scoring}.

\paragraph{AST Parser}

We observed that for the current code using undefined identifiers (\eg, \texttt{NameError} in Python), the infilling model typically succeeds in generating the corresponding missing statements. In such cases, we can employ the AST parser to \textit{deterministically} assess the correctness of the infill. The parser is able to analyze the AST of the current code, and identify any usage of undefined identifiers. If an infill $\mathbb{L}_i'$ correctly supplements this missing identifier (such as referencing an external library or defining a new function), we can ascertain the accuracy of $\mathbb{L}_i'$. In this case, we record the score of this infill as infinity, \ie, $V_i=+\infty$, forcing it to be utilized in the combination stage.

\paragraph{Generation Model Scoring}

While the AST parser can deterministically judge correct infills, it falls short in the cases where infills have latent utility but do not directly address undefined identifiers. For example, as illustrated in \cref{fig:motivation_example}, the AST parser cannot discern the infill (\texttt{open\_parens = 0}) due to the absence of subsequent references to this variable. To address this, a pivotal insight is that a trained LLM is adept at assessing the overall generation quality by computing token scores. We observed that \textit{correct infills usually lead LLMs to assign higher scores to subsequent tokens}, since they improve the overall code quality. Based on this finding, we introduce a scoring-based approach to determine the potential impact of infills.

Specifically, for the infilling line $\mathbb{L}_i'$ preceding the $i$-th line, we calculate the average token logit improvement of subsequent lines before and after combining $\mathbb{L}_i'$, namely:
\begin{align*}
    \Delta_k = \frac{1}{|L_k|} \sum_{j=1}^{|L_k|}\mathbb{S}'(x_{j}^{k}; \mathcal M_G) - S(x_{j}^{k}; \mathcal M_G),
\end{align*}
where $k\in\{i,\cdots,n\}$. $\mathbb{S}'(\cdot)$ and $S(\cdot)$ denotes the logit score determined by $\mathcal M_G$. $S(\cdot)$ is computed based on the initial code $[L_1\oplus \cdots\oplus L_n]$, and $\mathbb{S}'(\cdot)$ is computed based on the combined code $[L_1\oplus \cdots\oplus L_{i-1}\oplus \mathbb{L}_i'\oplus L_i\oplus \cdots\oplus L_n]$. $\Delta_k$ represents the improvement of line $L_k$ after combining $\mathbb{L}_i'$.

Next, we identify all consecutive lines after $\mathbb{L}_i'$ which exhibit notable improvement. We determine a $t$ (where $i\leq t \leq n$) such that the improvement for subsequent lines $\Delta_i, \Delta_{i+1}, \cdots, \Delta_t$ exceeds a pre-defined threshold $\tau$, and the improvement $\Delta_{t+1}$ for $L_{t+1}$ (if $n>t$) falls below $\tau$. If these improved lines are not more than half of all subsequent lines, \ie, $t-i+1\leq \nicefrac{(n-i+1)}{2}$, the infill is not considered beneficial for most of the subsequent lines and is thus disregarded. Otherwise, the total improvement $V_i$ from $L_i$ to $L_t$ is recorded as the score of this infill, formulated as $V_i=\sum_{k=i}^t \Delta_k$. Additionally, if $n>t$, we opt to \textit{remove} the extra lines after line $t$ when $\mathbb{L}_i'$ is combined, as they signify diminished improvement and necessitate revision.

\paragraph{Discussion}

Compared to AST Parser, Generation Model Scoring has a broader application scope. However, its judgments are not infallible and may occasionally result in errors. Tuning the improvement threshold $\tau$ can significantly reduce such misjudgments and increase the precision. We present examples of good and bad infills judged by \model~in \cref{sec:app_example} to clearly illustrate the role of Judging.

\subsection{Combination}\label{sec:method-combine}

Following the Judging stage, we filter out infills without significant improvement and record the score $V_i$ of each infill $\mathbb{L}_i'$. During the Combination stage, if all infills are filtered out, it indicates that no additional infill is necessary at the moment, and thus we adopt the standard generation line $\mathbb{L}_{n+1}'$. Otherwise, we adopt the infill with the highest score. 
\section{Experiments}

\subsection{Setup}

\paragraph{Models and Benchmarks}

We tested two open-source code LLMs: \textsc{CodeLlama} \citep{roziere_code_2023} and \textsc{WizardCoder-Python} \citep{luo_wizardcoder_2023}, with the latter being one of the SOTA models. \textsc{CodeLlama} was tested across 7B and 13B versions using the \textsc{Python} and \textsc{Instruct} variants, due to their superior performance in code generation compared to the base model. For \textsc{WizardCoder-Python}, we utilized the 13B and 34B versions. The infilling model used in our \model~is \textsc{CodeLlama-Instruct-7B} by default, which has been pre-trained with the FIM objective \citep{bavarian_efficient_2022}. 

We evaluated our approach across a range of widely-used code generation benchmarks, including HumanEval \citep{chen_evaluating_2021} and MBPP \citep{austin2021program}. To showcase the adaptability of our method across different programming languages, we also evaluate it on MultiPL-E \citep{cassano2022multipl}. For HumanEval and MBPP, we employed the prompts provided by \citet{bigcode-evaluation-harness} for \textsc{CodeLlama}'s evaluation and modified them to fit the instruction format of \textsc{WizardCoder} \citep{luo_wizardcoder_2023}. For MultiPL-E, following \citet{wei_magicoder_2023} we directly used the provided completion formats.




\begin{figure*}[t]
  \centering
  \subfigure[Java]{
  \label{fig:multiple_java}
    \includegraphics[width=0.31\textwidth,trim={5 5 5 5},clip]{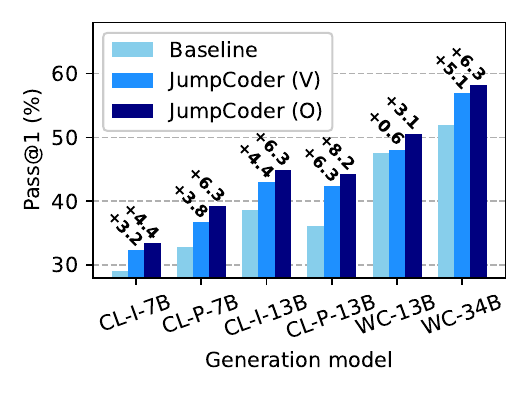}
  }
  \subfigure[C\#]{
  \label{fig:multiple_cs}
    \includegraphics[width=0.31\textwidth,trim={5 5 5 5},clip]{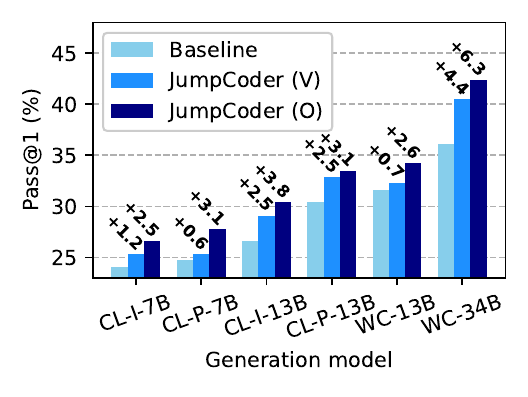}
  }
  \subfigure[C++]{
  \label{fig:multiple_cpp}
    \includegraphics[width=0.31\textwidth,trim={5 5 5 5},clip]{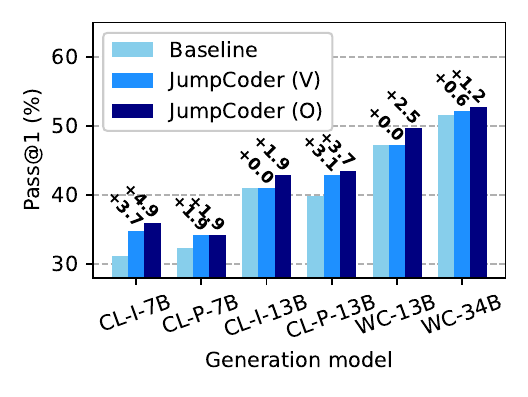}
  }
  \caption{Results on MultiPL-E. CL-I/P = \method{CodeLlama-Instruct/Python}. WC = \method{WizardCoder-Python}.}
  \label{fig:multiple}
\end{figure*}

\begin{table}[t!]
\centering
\small
\caption{Results of Pass@1 (\%) on HumanEval and MBPP using greedy generation. Following \citet{bigcode-evaluation-harness}, we use 0-shot for HumanEval and 1-shot for MBPP in our reproduced experiments. \textsc{JC} = \textsc{\underline{J}ump\underline{C}oder}. V = \underline{V}anilla. F = \underline{F}iltered. O = \underline{O}racle. $\dag$: Results are taken from \citet{roziere_code_2023}.}
\label{tab:main}

\resizebox{\linewidth}{!}{

\begin{tabular}{cccc}
\toprule
\textbf{Generation model} & \textbf{Method} & \textbf{HumanEval} & \textbf{MBPP} \\
\midrule
\textsc{GPT-3.5}$^\dag$ &      & 48.1 & 52.2 \\
\textsc{GPT-4}$^\dag$ &      & 67.0 & - \\
\textsc{StarCoder}$^\dag$ (15B) &      & 33.6 & 52.7 \\
\textsc{CodeLlama}$^\dag$ (7B) &      & 33.5 & 41.4 \\
\textsc{CodeLlama}$^\dag$ (13B) &      & 36.0 & 47.0 \\
\textsc{CodeLlama}$^\dag$ (34B) &      & 48.8 & 55.0 \\
\midrule
\multicolumn{1}{c}{\multirowcell{4}{\textsc{CodeLlama}\\\textsc{-Instruct}\\(7B)}} & -    & 36.0 & 42.4 \\
     & + \textsc{JC}~(V) & 37.8 (+1.8) & 44.8 (+2.4) \\
     & + \textsc{JC}~(F) & \textbf{39.6 (+3.6)} & \textbf{45.2 (+2.8)} \\
     & + \textsc{JC}~(O) & \textcolor[rgb]{ .502,  .502,  .502}{\textit{39.6 (+3.6)}} & \textcolor[rgb]{ .502,  .502,  .502}{\textit{45.2 (+2.8)}} \\
\arrayrulecolor{gray!80}
\midrule
\arrayrulecolor{black}
\multicolumn{1}{c}{\multirowcell{4}{\textsc{CodeLlama}\\\textsc{-Python}\\(7B)}} & -    & 38.4 & 43.2 \\
     & + \textsc{JC}~(V) & 40.2 (+1.8) & 45.4 (+2.2) \\
     & + \textsc{JC}~(F) & \textbf{41.5 (+3.1)} & \textbf{45.6 (+2.4)} \\
     & + \textsc{JC}~(O) & \textcolor[rgb]{ .502,  .502,  .502}{\textit{41.5 (+3.1)}} & \textcolor[rgb]{ .502,  .502,  .502}{\textit{46.8 (+3.6)}} \\
\arrayrulecolor{gray!80}
\midrule
\arrayrulecolor{black}
\multicolumn{1}{c}{\multirowcell{4}{\textsc{CodeLlama}\\\textsc{-Instruct}\\(13B)}} & -    & 40.9 & 45.8 \\
     & + \textsc{JC}~(V) & \textbf{44.5 (+3.6)} & \textbf{46.8 (+1.0)} \\
     & + \textsc{JC}~(F) & 43.9 (+3.0) & 46.6 (+0.8) \\
     & + \textsc{JC}~(O) & \textcolor[rgb]{ .502,  .502,  .502}{\textit{45.7 (+4.8)}} & \textcolor[rgb]{ .502,  .502,  .502}{\textit{48.0 (+2.2)}} \\
\arrayrulecolor{gray!80}
\midrule
\arrayrulecolor{black}
\multicolumn{1}{c}{\multirowcell{4}{\textsc{CodeLlama}\\\textsc{-Python}\\(13B)}} & -    & 43.9 & 50.0 \\
     & + \textsc{JC}~(V) & \textbf{45.7 (+1.8)} & \textbf{51.0 (+1.0)} \\
     & + \textsc{JC}~(F) & \textbf{45.7 (+1.8)} & 50.8 (+0.8) \\
     & + \textsc{JC}~(O) & \textcolor[rgb]{ .502,  .502,  .502}{\textit{47.0 (+3.1)}} & \textcolor[rgb]{ .502,  .502,  .502}{\textit{53.2 (+3.2)}} \\
\arrayrulecolor{gray!80}
\midrule
\arrayrulecolor{black}
\multicolumn{1}{c}{\multirowcell{4}{\textsc{WizardCoder}\\\textsc{-Python}\\(13B)}} & -    & 64.0 & 56.8 \\
     & + \textsc{JC}~(V) & 64.6 (+0.6) & \textbf{57.2 (+0.4)} \\
     & + \textsc{JC}~(F) & \textbf{65.2 (+1.2)} & \textbf{57.2 (+0.4)} \\
     & + \textsc{JC}~(O) & \textcolor[rgb]{ .502,  .502,  .502}{\textit{65.9 (+1.9)}} & \textcolor[rgb]{ .502,  .502,  .502}{\textit{57.2 (+0.4)}} \\
\arrayrulecolor{gray!80}
\midrule
\arrayrulecolor{black}
\multicolumn{1}{c}{\multirowcell{4}{\textsc{WizardCoder}\\\textsc{-Python}\\(34B)}} & -    & 73.8 & 59.2 \\
     & + \textsc{JC}~(V) & \textbf{74.4 (+0.6)} & 59.2 (+0.0) \\
     & + \textsc{JC}~(F) & \textbf{74.4 (+0.6)} & \textbf{59.6 (+0.4)} \\
     & + \textsc{JC}~(O) & \textcolor[rgb]{ .502,  .502,  .502}{\textit{75.0 (+1.2)}} & \textcolor[rgb]{ .502,  .502,  .502}{\textit{60.0 (+0.8)}} \\
\bottomrule
\end{tabular}%

}

\end{table}

\paragraph{Implementation Details}

In alignment with previous studies \citep{chen2023teaching,wei_magicoder_2023}, we applied greedy decoding for both generation and infilling in each task, and reported the Pass@1 metric. Besides, we found that \model~may degenerate the code in a few cases when the autoregressive generation is already correct. To further investigate these degenerated cases, we design two strategies for selecting one code from the codes generated by \model~(denoted as A) and traditional autoregressive coder (denoted as B): (1) \textbf{\model~(filtered)}, in which if all infills are judged by AST Parser instead of Generation Model Scoring, A is chosen, otherwise the code with lower perplexity (PPL) prevails; and (2) \textbf{\model~(oracle)}, comparing A and B with the test cases used for evaluation and selecting the better one, serving as the performance upper bound. We provide further implementation details in \cref{sec:app_jumpcoder_detail}.

\subsection{Results on HumanEval and MBPP}

\begin{figure}[t]
  \centering
  \includegraphics[width=0.47\textwidth,trim={10 10 10 10},clip]{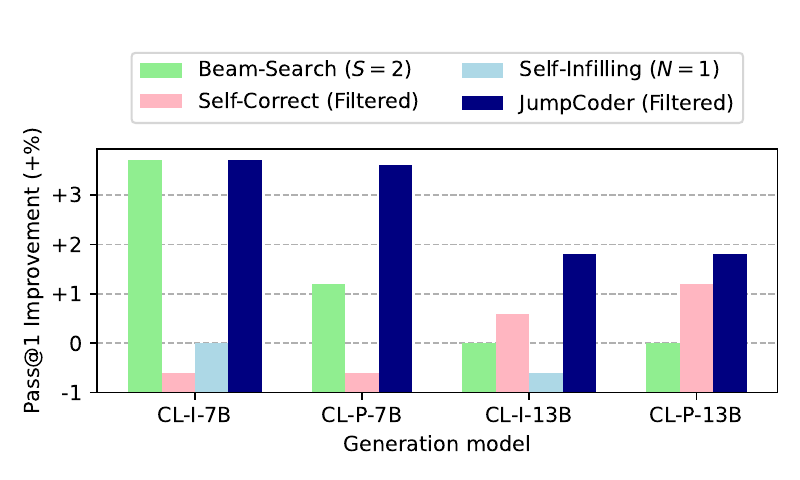}
  \caption{Performance improvements of various baselines over greedy decoding on HumanEval.}
  \label{fig:baselines}
\end{figure}

\cref{tab:main} summarizes results of Pass@1 on HumanEval and MBPP across various generation models. We can observe that \model~consistently enhances performance across both benchmarks compared to the base models. Specifically,

\begin{itemize}[leftmargin=*]
    \item \model~(Vanilla) shows a notable improvement over the baselines. It indicates that \model~can effectively improve existing code LLMs by addressing the irreversibility limitation. 
    \item \model's improvements are more pronounced with weaker baselines (\method{CodeLlama}) than with the stronger ones (\method{WizardCoder}), likely due to the former's challenges in establishing necessary foundations, \eg, predefined variables and references, for subsequent code (see Appendix \ref{sec:app_example} for concise examples). However, as discussed in the following section, \method{WizardCoder}'s capability for addressing these challenges does not stand out in specific languages.
    \item \model~(Vanilla) does not achieve the optimal potential, as indicated by \model~(Oracle) which is the performance upper bound, suggesting the presence of certain suboptimal infills. Such cases result in a decline from the original autoregressive coder. This issue is partially addressed by \model~(Filtered), which consistently outperforms \model~(Vanilla) and even reaches the performance of \model~(Oracle) in several scenarios.
\end{itemize}

\begin{figure*}[t]
  \centering
  \subfigure[Ablation study]{
  \label{fig:ablation}
    \includegraphics[width=0.31\textwidth,trim={5 5 5 5},clip]{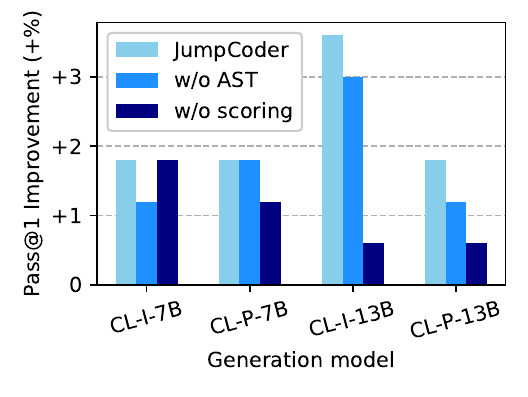}
  }
  \subfigure[Types of errors addressed]{
  \label{fig:error_type}
    \includegraphics[width=0.31\textwidth,trim={5 5 5 5},clip]{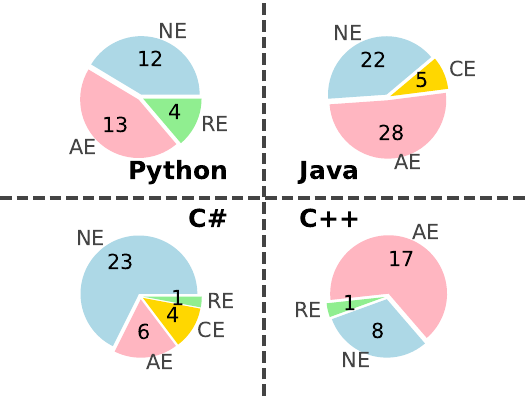}
  }
  \subfigure[Number of undefined identifiers]{
    \label{fig:name_error}
    \includegraphics[width=0.31\textwidth,trim={5 5 5 5},clip]{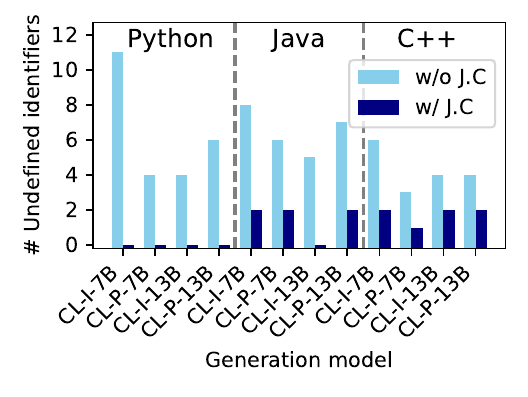}
  }
  \caption{\textbf{(a)} Ablation study of removing AST Parser and Generation Model Scoring during the judging stage.  \textbf{(b)} Different types of errors addressed by \model. NE = Name Error (\ie, undefined identifier error). AE = Assertion Error (\ie, failed at test cases). RE = Runtime Error. CE = Compile Error. \textbf{(c)} The number of undefined identifier errors for different languages. CL-I/P = \method{CodeLlama-Instruct/Python}.}
  \label{fig:addtional_experiment}
\end{figure*}

\begin{figure*}[htb]
  \centering
  \subfigure[Infilling model]{
  \label{fig:hyper-parameters-infill}
    \includegraphics[width=0.31\textwidth,trim={10 10 10 10},clip]{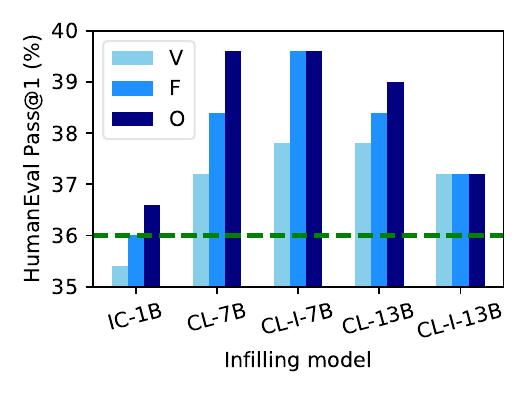}
  }
  \subfigure[Improvement threshold $\tau$]{
  \label{fig:hyper-parameters-tau}
    \includegraphics[width=0.31\textwidth,trim={10 10 10 10},clip]{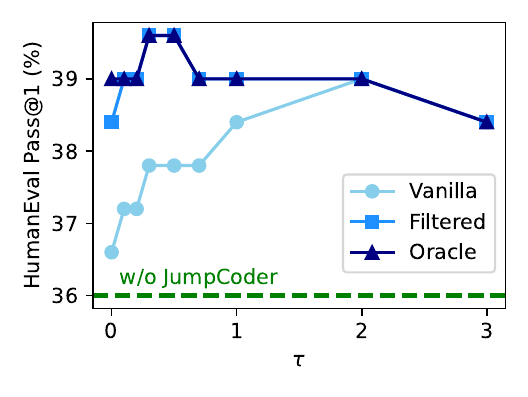}
  }
  \subfigure[Infill at top-$k$ positions]{
  \label{fig:hyper-parameters-k}
    \includegraphics[width=0.31\textwidth,trim={10 10 10 10},clip]{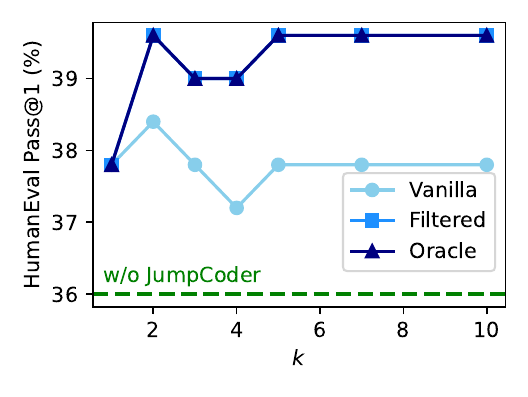}
  }
  \caption{Impact of infilling models and hyperparameters on three variants of \model~on HumanEval. The generation model is \method{CodeLlama-Instruct-7B}, whose performance is marked in the \textcolor[HTML]{008000}{green} dashed line. IC = \method{InCoder}. CL = \method{CodeLlama}. CL-I = \method{CodeLlama-Instruct}.}
  \label{fig:hyper-parameters}
\end{figure*}

\subsection{Results on MultiPL-E}

\cref{fig:multiple} illustrates the improvements achieved by \model~in Java, C\# and C++ using the MultiPL-E \citep{cassano2022multipl} benchmark. The findings reveal consistent enhancements across these programming languages. On average, \model~passes an additional 2.9\% (Python), 5.8\% (Java), 3.6\% (C\#) and 2.7\% (C++) problems. 
Remarkably, while \model's advancements over \method{WizardCoder} are less marked in Python, considerable gains are evident in Java and C\#. This implies the impact of irreversibility limitation may be language-dependent: certain languages pose more challenges for models not only in preemptively generating essential declarations or statements but also in producing suitable subsequent codes to compensate for previous omissions.


\subsection{Results Compared with Other Baselines}

In addition, we compared \model~with three similar baselines: \textsc{Self-Correct} \cite{huang2023large,madaan2023self,ganguli2023capacity}, \textsc{Beam-Search} \cite{DBLP:conf/emnlp/WisemanR16} and \textsc{Self-Infilling} \cite{zheng_self-infilling_2023}. Details on the baselines and the evaluation protocol are provided in \cref{sec:app_baselines}. \cref{fig:baselines} presents the evaluation results, which reveals \model's consistent superiority. Notably, \method{Beam-Search} outperforms \method{Greedy-Decoding} only with the 7B models, suggesting the 13B model's ability to generate correct code on the first try. In contrast, \method{Self-Correct} underperforms \method{Greedy-Decoding} with the 7B model, likely due to the limited capability of the small-size model in comprehending complicated instructions effectively. We also observed that \method{Self-Infilling} did not perform notably well, likely due to two factors: (1) The infilling capability of the instruct models may have been weakened due to the instruction fine-tuning \citep{zheng_self-infilling_2023}; (2) The effectiveness of \method{Self-Infilling} largely depends on rectifying empty code outputs (\eg, outputting placeholder statements like "\texttt{pass}"), an effect that is less evident in our setting where all the generated code has substantial content (refer to \cref{sec:app_baselines}).

\section{Analysis}

\paragraph{Ablation Studies} we evaluated \model~(Vanilla) by excluding the AST Parser (labeled as w/o AST) and Generation Model Scoring (labeled as w/o scoring). \cref{fig:ablation} illustrates the performance improvements of these variants compared to the complete \model~on various generation models. The results indicate both variants were less effective than the full \model. This is attributed to the complementary roles of the AST Parser and Generation Model Scoring: the AST Parser deterministically accepts the infills that resolve undefined identifier errors, which can be probabilistically rejected by Generation Model Scoring. Conversely, Generation Model Scoring is able to accept more good infills. This synergy allows more effective infills, highlighting the enhanced performance achieved by integrating both Judging components in \model.

\paragraph{Different Types of Errors Addressed}

\cref{fig:error_type} illustrates different error types addressed by \model. Notably, the enhancements brought by \model~stem not only from supplementing missing declarations in the generated code (\ie, address undefined identifier errors), but also from inserting essential statements that facilitate the correctness (\ie, address other errors). We provide detailed examples illustrating how \model~tackles different error types in Appendix \ref{sec:app_example}.

\paragraph{Addressing Undefined Identifier Errors}

\cref{fig:name_error} demonstrates the reduction in undefined identifier errors across various languages with \model's application. Notably, it eradicates such errors in Python and significantly decreases them in C++ and Java. This demonstrates that allowing the model to jump back and re-encode previously generated parts enables undefined identifiers to be defined, confirming \model's effectiveness. We also noticed that C++ and Java errors are not completely resolved due to the infilling model's occasional inability to correctly fill in missing library applications and functions. Addressing this issue further is left for future work.


\paragraph{Infilling Model, Parameter and Efficiency Analysis}

We adjusted various infilling models and two hyperparameters (improvement threshold $\tau$ and number of infills $k$) to examine their impact on the model performance. We first evaluated \method{InCoder} \cite{DBLP:conf/iclr/FriedAL0WSZYZL23}, \method{CodeLlama} \cite{roziere_code_2023} and their instruction variants as the infilling models, all of which pre-trained with the FIM objective \citep{bavarian_efficient_2022}. Results in \cref{fig:hyper-parameters-infill} show that \method{CodeLlama-7B} outperforms even the larger 13B model, suggesting smaller models can effectively power \model. We also found that \method{InCoder} underperformed due to inadequate function and library completion capabilities. Additionally, we observed that the \method{CodeLlama-13B-Instruct} frequently generated statements with incorrect indentation. One possible explanation is that the instruct fine-tuning diminishes its infilling proficiency.

\cref{fig:hyper-parameters-tau} shows the results when tuning the improvement threshold $\tau$. It indicates that a low hyperparameter $\tau$ ($<0.3$) encourages \model~to adopt \textit{aggressive} infilling, which leads to the integration of numerous inferior infills, reducing overall performance. When $\tau$ is high ($>2$), \model's \textit{conservative} infill selection results in the exclusion of potentially beneficial infills, marginally diminishing performance. An intermediary $\tau$ enables the model to achieve an optimal trade-off between good and bad infills.

\cref{fig:hyper-parameters-k} shows the results on adjusting the maximum infill positions $k$. It suggests that a small $k$ limits \model's infill options, inadvertently filtering out advantageous infills and thereby decreasing performance. As $k$ increases, \model's performance reaches a plateau. Interestingly, \model~(Vanilla) exhibits an initial performance dip with an increase in $k$, likely due to the model selecting more suboptimal infills. Fortunately, \model~(Filtered) can effectively filter out these less desirable infills.

Additionally, we provide an in-depth discussion on generation speed and memory analysis in Appendix \ref{sec:generation_rate}, which reveals that \model's generation speed is $0.7\times$ that of autoregressive generation, maintaining a comparable order of magnitude.



\section{Related Work}

In this section, we review two lines of related work: code generation and code infilling.

\paragraph{Code Generation} 
Training large language models with an extensive corpus of code is a common approach for code generation tasks \cite{zan2023large}. There are several large language models trained on massive code data, such as Codex \cite{chen_evaluating_2021}, CodeGen \cite{DBLP:conf/iclr/NijkampPHTWZSX23}, AlphaCode \cite{li2022competition}, CodeGeeX \cite{zheng2023codegeex}, StarCoder \cite{li_starcoder_2023} and Code Llama \cite{roziere_code_2023}. These code generation models work in an autoregressive manner, which results in their inability to do online modification. Some studies focus on improving the work of autoregressive code generation models. In self-debugging \cite{chen2023teaching}, the model generates multiple iterations of code based on the explanation and execution results of the initially generated code. In contrast, our approach enables the model for online modification during a single round. Another concurrent study, self-infilling \cite{zheng_self-infilling_2023}, explores enhancing autoregressive code generation through infilling capabilities. Our approach differs from self-infilling in three key aspects: Firstly, self-infilling requires the generation model to possess infilling abilities and mandates specific suffix formats for different languages, which limits its applicability. Secondly, our method allows for online modifications, aligning more closely with human coding practices, whereas self-infilling merely alters the order of generation. Lastly, the two methods are somewhat complementary: certain self-infilling features, such as loops, can be integrated into our method, and our capabilities for online modification could also enhance their infilling or generation processes.

Several research targets improving LLM-based code generation \citep{DBLP:conf/iclr/ChenZNZLLC23,dong2023codep,le2022coderl,DBLP:conf/iclr/FriedAL0WSZYZL23,shen2023pangu}. CodeT \cite{DBLP:conf/iclr/ChenZNZLLC23} utilizes one language model to generate both code snippets and test cases. Subsequently, it ranks the code snippets by assessing the consistency between the code snippets and the associated test cases. CODEP \cite{dong2023codep} enhances the existing code generation framework by incorporating an automatic pushdown automaton (PDA) module to ensure the syntactic correctness of generated code. Recently, many code generation approaches fine-tuned code LLMs with high-quality instruction fine-tuning datasets \citep{luo_wizardcoder_2023,wei_magicoder_2023,gunasekar2023textbooks,muennighoff2023octopack}. Our approach also harmonizes with these models and can be used to enhance their performance.

\paragraph{Code Infilling}\label{sec:code_infilling}
Various code generation models possess the capability of code infilling. While excelling in code generation, decoder-only architecture code generation models face limitations in understanding context due to their unidirectional attention mechanism. In response to this architectural limitation, these models employ methodologies akin to the FIM \cite{bavarian_efficient_2022} concept. The model learns to fill these regions in a standard left-to-right autoregressive manner, acquiring infilling capabilities. This approach has been used to pre-train on various code generation models, including Incoder \cite{DBLP:conf/iclr/FriedAL0WSZYZL23}, StarCoder \cite{li_starcoder_2023}, CodeGen 2 \cite{nijkamp2023codegen2} and CodeLlama \cite{roziere_code_2023}, and has boosted downstream tasks such as code completion and document generation. In this work, we leverage these code-infilling capabilities to enable online modification and enhance code generation.



\section{Conclusion}
This research analyzes the inherent irreversibility limitation of the autoregressive sequential generation, a critical yet underexplored issue. We introduce \model, a framework that enables human-like online code modification for existing LLMs through an infilling model without retraining. This framework is readily adaptable to diverse programming languages. Through extensive experimentation, we demonstrate that our proposed \model~significantly enhances code generation quality across various code LLMs and benchmarks. We hope our work can motivate more future investigations into this limitation of LLMs.

\section*{Limitations}


There are some worthwhile directions for future research to address the limitations in this paper, which we list below:

\begin{itemize}[leftmargin=*]
    \item Further optimizing the time/memory efficiency of \model. In our framework, we employ an \textit{infill-first, judge-later} strategy. While parallel generation and speculative infilling are utilized to expedite the process, it remains comparatively slower than traditional generation methods and increases memory consumption. Future research could focus on exploring alternative paradigms for online modification during generation, such as devising a component to preemptively identify the most effective infill positions.
    \item Refine the judging process for infills. While the current judging approach proves effective in our experiments, it depends on heuristic rules and may sometimes endorse suboptimal infills. Developing a more generalized judging framework to enhance the synergy between generation and infilling models presents a compelling research avenue.
    \item Generalize \model's capabilities and scope. One limitation of \model~is that it can only add new code within the current code. Broadening the online modification to encompass online deletions or changes offers a compelling expansion path. Additionally, the challenge of irreversibility is intrinsic to autoregressive generation across various applications, not exclusively to code generation. Despite being tailored for code generation, the promising results from our framework hint at the broader utility for tasks such as mathematical problem-solving and natural language generation.
    
\end{itemize}

\section*{Ethics Statement}

Our work complies with the ACL Ethics Policy. All datasets and models are publicly accessible. We have not identified any significant ethical considerations associated with our work. We believe our findings can inspire further research into the irreversibility limitation of LLMs.

\section*{Acknowledgments}

This research is supported by the National Natural Science Foundation of China (No. 62202420) and the Software Engineering Application Technology Lab at Huawei under the Contract TC20231108060. Zhongxin Liu gratefully acknowledges the support of Zhejiang University Education Foundation Qizhen Scholar Foundation.

\bibliography{ref}

\appendix

\newpage

\numberwithin{equation}{section}

\section*{Appendix}

\section{Algorithm}\label{sec:app_algorithm}

In this section, we present pseudocode diagrams of our method to facilitate a clearer understanding of the operational workflow. \cref{alg:Hybrid generation Algorithm} involves generating the next line of code or filling in the Top-$k$ critical lines in the current code based on the available code. \cref{alg:Judging Algorithm} assesses the validity of infills. \cref{alg:Combination Algorithm} selects the best infill based on the scores. Finally, \cref{alg:JumpCoder} outlines the workflow of our entire framework.

\SetAlgoNlRelativeSize{+0.5}
\SetKwInOut{Input}{Input}
\SetKwInOut{Output}{Output}
\SetKw{Continue}{continue}
\SetKwComment{Comment}{\color{red}$\triangleright$\ \tiny}{}
\SetKw{Break}{break}
\algnewcommand\algorithmicparallel{\textbf{parallel}}
\algdef{S}[FOR]{ParallelFor}[1]{\algorithmicfor\ #1\ \algorithmicparallel}

\begin{algorithm}[!ht]
  \caption{Hybrid generation Algorithm of \model}
  \label{alg:Hybrid generation Algorithm}
  \small
 \Input{Lines of current code $L$, generation model $\mathcal M_G$, infilling model $\mathcal M_I$, hyperparameters $k$. }
 \Output{Generated / infilled lines in each position $\mathbb{L}'$, infill positions $\mathcal I$.}
  $n\gets |L|$\;
  $\mathbb{L}' = \varnothing$\;
    \Comment{\color{red}Evaluate the first non-indent token score for each line.}
    \For{$i\gets1, \cdots, n$ \textnormal{\textbf{parallel}}}{
        $S_i\gets S(x_{\mathbf{0}}^i; \mathcal M_G)$\;
    }
    \Comment{\color{red}Get the critical positions based on top-$k$ lowest $S_i$.}
    $\mathcal I\gets {\arg \text{Top-}k}\ S_i$\;
    \Comment{\color{red}Infilling.}
    \For{$i\in \mathcal I$ \textnormal{\textbf{parallel}}}{
        $\mathbb{L}_i' \sim P_{\mathcal M_I}(\cdot \mid \texttt{<PRE>}\oplus L_{[1:i-1]}\oplus \texttt{<SUF>}\oplus L_{[i:n]})$\;
        $\mathbb{L}' \gets  \mathbb{L}'  \cup \{\mathbb{L}'_i\}$\;
    }
    \Comment{\color{red}Generation.}
    $\mathbb{L}'_{n+1}\sim P_{\mathcal M_G}(\cdot \mid L_{[1:n]})$\;
    $\mathbb{L}' \gets  \mathbb{L}'  \cup \{\mathbb{L}'_{n+1}\}$\;
\Return{$\mathbb{L}',\mathcal I$}
\end{algorithm}

\begin{algorithm}[!ht]
  \caption{Judging Algorithm of \model}
    \label{alg:Judging Algorithm}
 \small
 \Input{Lines of current code $L$, generated / infilled lines in each position $\mathbb{L}'$, generation model $\mathcal M_G$, hyperparameters $\tau$, infill positions $\mathcal I$. }
 \Output{Set of infill scores $\mathbb{V}$ and the corresponding updated codes $\mathbb{L}^{\text{next}}$.}
   $n\gets |L|$\;
   $\mathbb{L}^{\text{next}} = \varnothing$\;
   $\mathbb{V} = \varnothing$\;
    \For{$i\in \mathcal I$}{
        \Comment{\color{red}Judge $\mathbb{L}'_i$ by AST Parser.}
        \If{undefined identifier error is resolved}{
            \Comment{\color{red}We directly assign the highest score to this infill to force it to be accepted.}
            $V_i\gets+\infty$\;
            $L_i^{\text{next}}\gets L_1\oplus L_2\oplus \cdots\oplus L_{i-1}\oplus \mathbb{L}'_i\oplus L_{i}\oplus \cdots\oplus L_{n}$\;
            $\mathbb{V} \gets  \mathbb{V}  \cup \{V_i\}$\;
            $\mathbb{L}^{\text{next}} \gets  \mathbb{L}^{\text{next}}  \cup \{L_i^{\text{next}}\}$\;
            \Continue\;
        }
        \Comment{\color{red}Judge $\mathbb{L}'_i$ by Generation Model Scoring.}
        \Comment{\color{red}Find the last line ($t$) that have significant improvements.}
        $k\gets i - 1,\quad t\gets i - 1$\;
        \Repeat{$k + 1>n$ \textnormal{\textbf{ or }} $\Delta_k \leq \tau$}{
            $k\gets k+1$\;
            $\Delta_k \gets \frac{1}{|L_k|} \sum_{j=1}^{|L_k|}\mathbb{S}'(x_{j}^{k}; \mathcal M_G) - S(x_{j}^{k}; \mathcal M_G)$\;
            \If{$\Delta_k > \tau$}{
                $t\gets k$\;
            }
        }
        \If{$t-i+1\leq \nicefrac{(n-i+1)}{2}$}{
            \Comment{\color{red}$\mathbb{L}_i'$ is not beneficial for more than half subsequent lines.}
            \Comment{\color{red}Ignore it.}
            \Continue\;
        }
        \Comment{\color{red}Accept $\mathbb{L}_i'$ and record the average improvement.}
        $V_i\gets {\sum_{k=i}^t \Delta_k}$\;
        \Comment{\color{red}{Combine $\mathbb{L}_i'$ and remove the lines without improvement}.}
        $L_i^{\text{next}}\gets L_1\oplus L_2\oplus \cdots\oplus L_{i-1}\oplus \mathbb{L}'_i\oplus L_{i}\oplus \cdots\oplus L_{t}$\;
        $\mathbb{V} \gets  \mathbb{V}  \cup \{V_i\}$\;
        $\mathbb{L}^{\text{next}} \gets  \mathbb{L}^{\text{next}}  \cup \{L_i^{\text{next}}\}$\;
    }
\Return{$\mathbb{V},\mathbb{L}^{\text{next}}$}
\end{algorithm}

\begin{algorithm}[!ht]
  \caption{Combination Algorithm of \model}
  \label{alg:Combination Algorithm}
  \small
 \Input{Lines of current code $L$, set of infill scores $\mathbb{V}$ and the corresponding updated codes $\mathbb{L}^{\text{next}}$, generated line $\mathbb{L}'_{n+1}$. }
 \Output{Next iteration of code.}
    $n\gets |L|$\;
    \eIf{$\mathbb{V} = \varnothing$}{
        \Comment{\color{red}Good infill doesn't exist. We accept the line generated by generation model $\mathcal M_G$}
        \Return{$L_1\oplus L_2\oplus \cdots\oplus L_n\oplus \mathbb{L}'_{n+1}$}\;
    }{
        \Comment{\color{red}Find the infill with the highest value.}
        $i^*\gets \arg\max \mathbb{V}$\;
        \Return{$\mathbb{L}_{i^*}^{\text{next}}$}\;
    }

\end{algorithm}

\begin{algorithm}[!ht]
  \caption{\model}
  \label{alg:JumpCoder}
  \small
  \Input{Input prompt $\mathcal P$, generation model $\mathcal M_G$, infilling model $\mathcal M_I$, hyperparameters $\tau$ and $k$. }
  \Output{Lines of generated code $L$.}
  $L \gets \mathcal P$\;
  \While{the stopping criteria is not met}{
    \Comment{\color{red}Stage 1. Hybrid generation}
    Invoke hybrid generation (\cref{alg:Hybrid generation Algorithm}) with input $L, \mathcal M_G, \mathcal M_I, k$\;
    Obtain generated / infilled lines in each position $\mathbb{L}'$ and infill positions $\mathcal I$\;
    \Comment{\color{red}Stage 2. Judging}
    Invoke Judging (\cref{alg:Judging Algorithm}) with input $L, \mathbb{L}', \mathcal M_G, \tau, \mathcal I$\;
    Obtain set of infill scores $\mathbb{V}$ and the corresponding updated codes $\mathbb{L}^{\text{next}}$\;
    \Comment{\color{red}Stage 3. Combination}
    Invoke Combination (\cref{alg:Combination Algorithm}) with input $\mathbb{V}, \mathbb{L}^{\text{next}}, \mathbb{L}'$ and update $L$\;
  }
\Return{$L$}
\end{algorithm}


\section{Further Experimental Details and Results}\label{sec:app_addtional_exp}




\subsection{Implementation Details of \model}\label{sec:app_jumpcoder_detail}

\paragraph{Computation Sources} We conducted all the experiments in eight NVIDIA A800 GPUs for around 100 GPU hours. All models are loaded in the \texttt{float16} format.

\paragraph{Hyperparameters} Regarding our
method’s hyperparameters $k$ and $\tau$, by default we set $k = 5$ for Python and $k = 10$ for C++ and Java tasks. $\tau$ is set at 0.8 for C++, and 0.3 for other languages.

\paragraph{AST Parsers} The AST Parser is employed during the Judging stage to detect undefined identifiers in code snippets. For Python, we utilize the built-in \texttt{ast} module to analyze the AST of the current code. For C\#, we leverage \texttt{tree-sitter}\footnote{\url{https://github.com/tree-sitter/tree-sitter-c-sharp}} to parse the AST. For Java and C++, we invoke their respective compilers to compile the code and analyze error logs to determine the presence of undefined identifiers. Specifically, we identify undefined identifiers by checking if the compilation error logs contain \texttt{"cannot find symbol"} (for Java) and \texttt{"was not declared in this scope"} (for C++). To prevent syntax errors such as mismatched braces during incomplete generation, we ensured the balancing of braces first before invoking the compiler.

\paragraph{Tokenization} Tokenization in infilling models is challenging, as infilling at certain positions can disrupt token boundaries, leading to irregular tokens and degraded performance \cite{zheng_self-infilling_2023}. \model's infilling, strategically placed at the beginning of lines, aligns with the tokenization approaches of the infilling models we tested  (\method{CodeLlama} \cite{roziere_code_2023} and \method{InCoder} \cite{DBLP:conf/iclr/FriedAL0WSZYZL23}), which treat newline characters as distinct tokens (\texttt{[\textbackslash n]}). This placement minimizes the incidence of irregular tokens. However, it's noteworthy that some tokenizers, like StarCoder \cite{li_starcoder_2023}, combine \texttt{[\textbackslash n]} with subsequent indentation into a single token. It renders our current infilling strategy inappropriate for such tokenizers since this tokenizer strategy breaks the token. Addressing this compatibility issue comprehensively is earmarked for future research.

\paragraph{Multi-line Function Infills} Given \model's line-level operation, our approach to infilling is limited to single-line infilling. A straightforward method is employing the newline token \texttt{[\textbackslash n]} as a stop token in the infilling stage, substantially boosting the process's efficiency. However, we also recognized the need for multi-line continuity when filling the missing functions. Consequently, we implemented an adaptive stop criterion: the newline token serves as the default stop token. If the generation of a function signature is detected, this stop token is eliminated, facilitating the function's completion. This strategy effectively harmonizes the efficiency of line-level and function-level infilling, maintaining simplicity and clarity.

\paragraph{Filter Repetition of Infills} It is known that language models tend to repeat previous sentences \citep{xu2022learning,radford2019language,keskar2019ctrl}, leading infilling models to occasionally produce contextually redundant statements. These poorly suited infills, frequently receiving higher scores from generation models, are often accepted during Judging. To address this, we utilized the gestalt pattern matching algorithm \citep{ratcliff1988pattern} to exclude the infills that are overly similar to the existing context, applying a similarity threshold of 0.85.

\paragraph{Filter Invalid Multi-line Block Infills} As \model~is currently restricted to single-line infills (except for function-level infilling), generating initial lines of code blocks like \texttt{if/for/while/try} often results in syntactic inconsistencies with the subsequent code. Therefore, we exclude infills that begin with keywords such as \texttt{if/for/while/try}, deferring the infilling of entire code blocks to future work for enhanced coherence and syntactic compatibility.

\subsection{Implementation Details of Baselines and Evaluation Protocol}\label{sec:app_baselines}

In this section, we introduce several similar baselines and compare them with~\model. We selected the following baseline:


\paragraph{Beam-Search and Rank} \method{Beam-Search} \cite{DBLP:conf/emnlp/WisemanR16} manages distinct search beams and makes a better selection during the decoding. Given that \model~requires generating up to $k=5$ infills at each iteration, we demonstrate the benefits of this variant by comparing it with autoregressive generation five times, specifically using \textsc{Beam-Search} with a beam size of 5. We also compare \textsc{Rank} \cite{chen_evaluating_2021}, which generates $k=5$ codes and ranks them with the mean log probability. We used nucleus sampling with $p=0.95$ and a temperature of 0.8, following the setting in \citet{chen_evaluating_2021}.

\paragraph{Self-Infilling} Similar to our use of an infilling model for online modification, \method{Self-Infilling} \cite{zheng_self-infilling_2023} employs infilling to augment generation processes. However, its application necessitates inherent infilling capabilities within the model, restricting its use to \method{CodeLlama-Instruct} models in our studies. We tested the recommended $N=2$ setting where $N$ denotes the number of the loop times within \method{Self-Infilling}. Since it was verified in their paper that the looping mechanism strategy is important, we report their performance using both two looping mechanism strategies, \textit{Vanilla Split}, and \textit{Extended Split}.


\paragraph{Self-Correct} Self-Correct \cite{huang2023large,madaan2023self,ganguli2023capacity} allows the model to generate two rounds and the second round aims to correct the code generated in the first round. It is akin to our approach of online modification, although its correction is offline. We used the prompt similar to \citet{huang2023large}:

\definecolor{lightgray}{gray}{0.9}

\lstdefinestyle{autoregressioncoderstyle}{
  basicstyle=\fontsize{9}{11}\ttfamily,
  keywordstyle=\color{blue}, 
  commentstyle=\color{gray}\bfseries,
  morekeywords={if,else,while}, 
  moredelim=[is][\bfseries]{@}{@}, 
  moredelim=[is][\color{black}]{*}{*}, 
  escapeinside={(*@}{@*)}, 
}
\begin{minipage}[t][5.5cm]{0.45\textwidth}
\begin{lstlisting}[style=autoregressioncoderstyle]
@{Problem Prompt}@
*{Code generated in the first round}*

"""
Review the above code and find problems (e.g., NameError) with the code. Based on the problems you found, improve your answer. Below is the refined code:
"""

@{Problem Prompt}@
*{Code generated in the second round}*
\end{lstlisting}
\end{minipage}

Similar to the variants used in our \model, we refer to the code generated in the second round as \textbf{Vanilla}, name the code from the two rounds with the lower PPL as \textbf{Filtered}, and label the code from the two rounds that performs better in the evaluation stage as \textbf{Oracle}.

We also noticed that some recent research has created enhanced prompts to incorporate the compiler feedback after running test cases \citep{chen2023teaching,zhang2023selfedit,zhou2023language} or human feedback \citep{wang2023chatcoder} for improved reasoning and coding. However, these approaches significantly alter the prompt and introduce considerable extra information to enhance LLMs, which is out of the scope of this paper. We choose the basic \method{Self-Correct} as our focus is fairly comparing our online modification approach with the offline self-correct approach. Besides, such research concerning advanced prompts is complementary to our approach. For instance, our method can be applied in the second (or subsequent) round of generation with these baselines. Exploring the integration of powerful prompt techniques with our method is an avenue for future research.

\begin{table*}[htbp]
\centering
\small
\caption{(Full results of \cref{fig:baselines}) The comparative analysis of \model~against various baselines.}
\label{tab:baseline}
\begin{scriptsize}

\begin{tabular}{lrcccc}
\toprule
\multirow{2}[4]{*}{\textbf{Generation model}} & \multirow{2}[4]{*}{\textbf{Size}} & \multirow{2}[4]{*}{\textbf{Method}} & \multicolumn{3}{c}{\textbf{Setting}} \\
\cmidrule{4-6}     &      &      & Vanilla & Filtered & Oracle \\
\midrule
\multirow{7}[2]{*}{\textsc{CodeLlama-Instruct}} & \multirow{7}[2]{*}{7B} & \method{Greedy-Decoding} & 37.8 & -    & - \\
     &      & \method{Beam-Search} $(S=5)$ & 37.2    & - & - \\
     &      & \method{Self-Infilling} (\textit{Vanilla split}, $N=2$) & 36.0    & -  & - \\
     &      & \method{Self-Infilling} (\textit{Extended split}, $N=2$) & 29.3    & -  & - \\
     &      & \method{Rank} $(k=5)$ & 39.0   & -  & - \\
     &      & \method{Self-Correct} & 37.2 & 37.2 & 39.0 \\
     &      & \model & \textbf{40.2} & \textbf{41.5} & \textbf{41.5} \\
\arrayrulecolor{gray!80}
\midrule
\arrayrulecolor{black}
\multirow{4}[2]{*}{\textsc{CodeLlama-Python}} & \multirow{4}[2]{*}{7B} & \method{Greedy-Decoding} & 42.7 & -    & - \\
     &      & \method{Beam-Search} $(S=5)$ & 41.4     & -  & - \\
     &      & \method{Rank} $(k=5)$ & 43.9   & -  & - \\
     &      & \method{Self-Correct} & 42.1  & 42.1  & 42.7 \\
     &      & \model & \textbf{46.3} & \textbf{46.3} & \textbf{47.6} \\
\arrayrulecolor{gray!80}
\midrule
\arrayrulecolor{black}
\multirow{7}[2]{*}{\textsc{CodeLlama-Instruct}} & \multirow{7}[2]{*}{13B} & \method{Greedy-Decoding} & 43.9 & -    & - \\
     &      & \method{Beam-Search} $(S=5)$ & 43.9  & -  & - \\
     &      & \method{Self-Infilling} (\textit{Vanilla split}, $N=2$) & 37.2    & -  & - \\
     &      & \method{Self-Infilling} (\textit{Extended split}, $N=2$) & 36.0    & -  & - \\
     &      & \method{Rank} $(k=5)$ & 42.7   & -  & - \\
     &      & \method{Self-Correct} & 44.5 & 44.5 & 44.5  \\
     &      & \model & \textbf{45.7} & \textbf{45.7} & \textbf{48.2} \\
\arrayrulecolor{gray!80}
\midrule
\arrayrulecolor{black}
\multirow{4}[2]{*}{\textsc{CodeLlama-Python}} & \multirow{4}[2]{*}{13B} & \method{Greedy-Decoding} & 48.8 & -    & - \\
     &      & \method{Beam-Search} $(S=5)$ & 49.3    & -  & - \\
     &      & \method{Rank} $(k=5)$ & 50.0   & -  & - \\
     &      & \method{Self-Correct} & 50.0  & 50.0 & 50.6 \\
     &      & \model & \textbf{50.6} & \textbf{50.6} & \textbf{51.2} \\
\bottomrule
\end{tabular}%

\end{scriptsize}

\end{table*}

\paragraph{Evaluation Protocol}

We found that existing code LLMs often perform degenerate behaviors resulting in empty program outputs, like mere \texttt{pass} or \texttt{"\# TODO: implement me"} statements. This significantly obstructs the generation of valid programs and hurts the performance. To ensure a focused comparison of the performance of valid programs, we restricted the model from generating tokens \texttt{\#} and \texttt{pass}. This strategy eliminates such invalid placeholder outputs, resulting in valid programs for all problems and a notable improvement in the performance of both the base models and our method.


\cref{tab:baseline} summarizes the full results, which shows that our \model~outperforms the baselines across different models and settings. 

\subsection{Efficiency Comparison}\label{sec:generation_rate}



\begin{figure}[t!]
  \centering
  \includegraphics[width=0.31\textwidth,trim={10 10 10 10},clip]{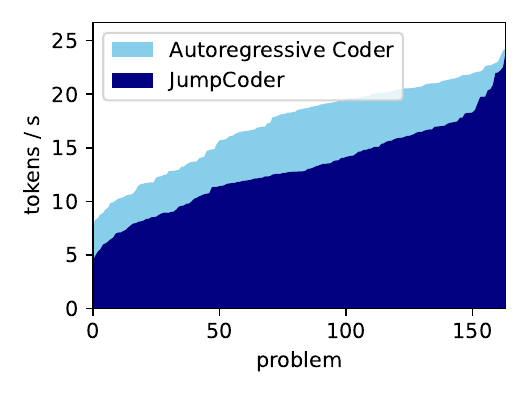}
  \caption{Number of tokens generated per second of the \model~on HumanEval for each problem. The generation model utilized is \method{CodeLlama-Instruct-7B}, with \method{CodeLlama-7B} being the infilling model. The number of infill positions ($k$) is set to 5. Problems are sorted according to their generation speed.}
  \label{fig:exp_run_time}
\end{figure}

\paragraph{Speed comparison} \cref{fig:exp_run_time} shows the evaluation results on the efficiency comparison of \model~and traditional autoregressive coder. The average generation speed of \textsc{JumpCoder} is about $0.7\times$ that of autoregressive generation. This demonstrates that \model's generation speed remains within the same order of magnitude as autoregressive generation. Furthermore, it indicates that \model~avoids significant performance degradation across different problems.

\begin{figure}[t!]
  \centering
  \includegraphics[width=0.31\textwidth,trim={10 10 10 10},clip]{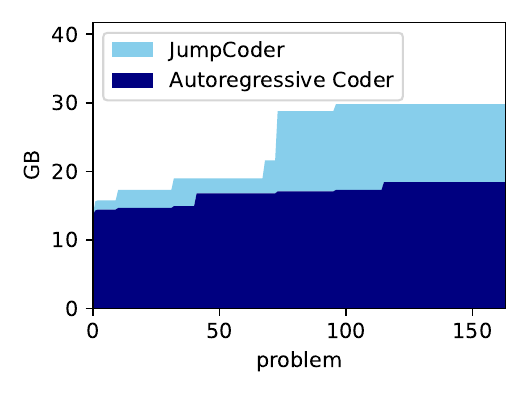}
  \caption{Memory usage of \model~on HumanEval when both the generation model and infilling model are the same \method{CodeLlama-Instruct-7B} for each problem. The number of infill positions ($k$) is set to 5. Problems are sorted according to the memory usage.}
  \label{fig:exp_memory}
\end{figure}

\paragraph{Memory comparison}  Our \model~can utilize the same model for both generation and infilling tasks, or employ two separate models. To compare the runtime overhead between \model~and autoregressive coder, we use a single 7B model as both models, which avoids the inclusion of additional static model parameter storage in the comparison. \cref{fig:exp_memory} displays the runtime memory overhead for \model. We find that \model~introduces up to nearly double the usage overhead, due to it potentially generating up to $k=5$ infills simultaneously at maximum. Fortunately, with our speculative infilling optimization, many problems do not require generating so many infills simultaneously, which saves a part of runtime GPU memory.

\section{Examples of \model}\label{sec:app_example} 

\begin{figure*}[!htb]
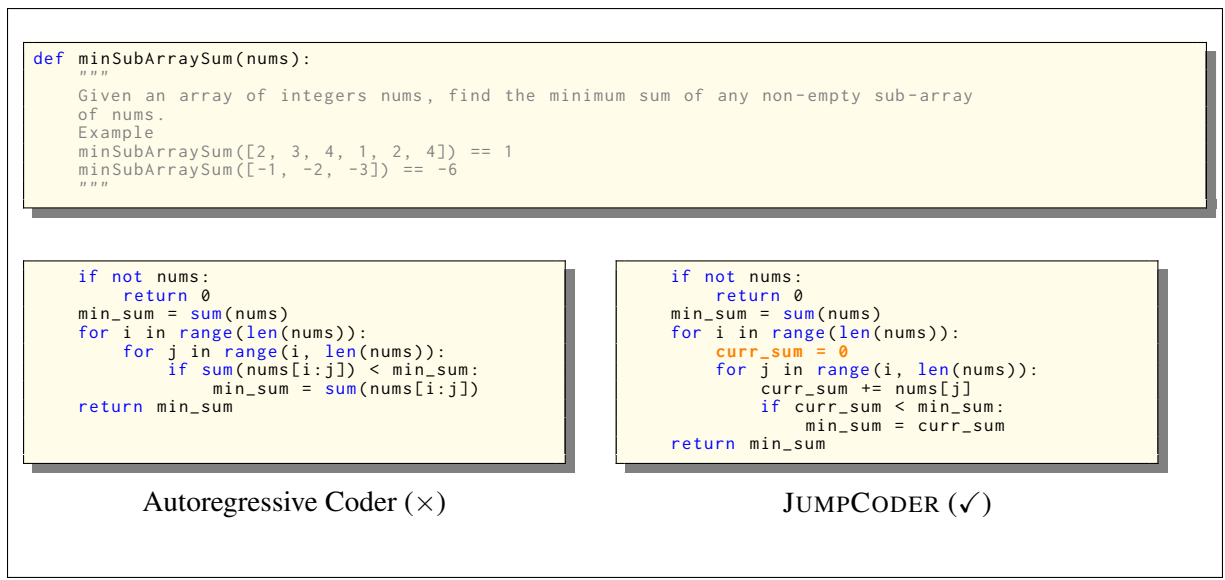

\begin{framed}

\begin{lstlisting}
def minSubArraySum(nums):
    """
    Given an array of integers nums, find the minimum sum of any non-empty sub-array
    of nums.
    Example
    minSubArraySum([2, 3, 4, 1, 2, 4]) == 1
    minSubArraySum([-1, -2, -3]) == -6
    """
\end{lstlisting}

\lstdefinestyle{autoregressioncoderstyle}{
  basicstyle=\fontsize{7}{7}\ttfamily,
  keywordstyle=\color{blue}, 
  morekeywords={if,else,while}, 
  framexbottommargin=14pt, 
  moredelim=[is][\color{red}\bfseries]{@}{@}, 
  escapeinside={(*@}{@*)}, 
}
\lstdefinestyle{jumpcoderstyle}{
  basicstyle=\fontsize{7}{7}\ttfamily,
  keywordstyle=\color{blue}, 
  morekeywords={if,else,while}, 
  moredelim=[is][\color{green!70!black}\bfseries]{@}{@}, 
  moredelim=[is][\color{orange}\bfseries]{*}{*}, 
}
\begin{minipage}[t][4cm]{0.45\columnwidth}
\begin{lstlisting}[style=autoregressioncoderstyle]
    if not nums:
        return 0
    min_sum = sum(nums)
    for i in range(len(nums)):
        for j in range(i, len(nums)):
            if sum(nums[i:j]) < min_sum:
                min_sum = sum(nums[i:j])
    return min_sum
\end{lstlisting}
\centerline{Autoregressive Coder ($\times$)}
\end{minipage}
\hspace{20pt}
\begin{minipage}[t][4cm]{0.45\columnwidth}
\begin{lstlisting}[style=jumpcoderstyle]
    if not nums:
        return 0
    min_sum = sum(nums)
    for i in range(len(nums)):
*        curr_sum = 0* 
        for j in range(i, len(nums)):
            curr_sum += nums[j]
            if curr_sum < min_sum:
                min_sum = curr_sum
    return min_sum
\end{lstlisting}
\centerline{\model~(\checkmark)}
\end{minipage}

\end{framed}
\caption{An example program: In this example, our approach prompts the model to fill in a variable for tracking the current \texttt{sum(subarray)}, while the autoregressive coder uses the \texttt{sum} function for summation. This helps avoid errors arising from the use of list slicing operations.}
\label{fig:good_example1}
\end{figure*}

\begin{figure*}[!htb]
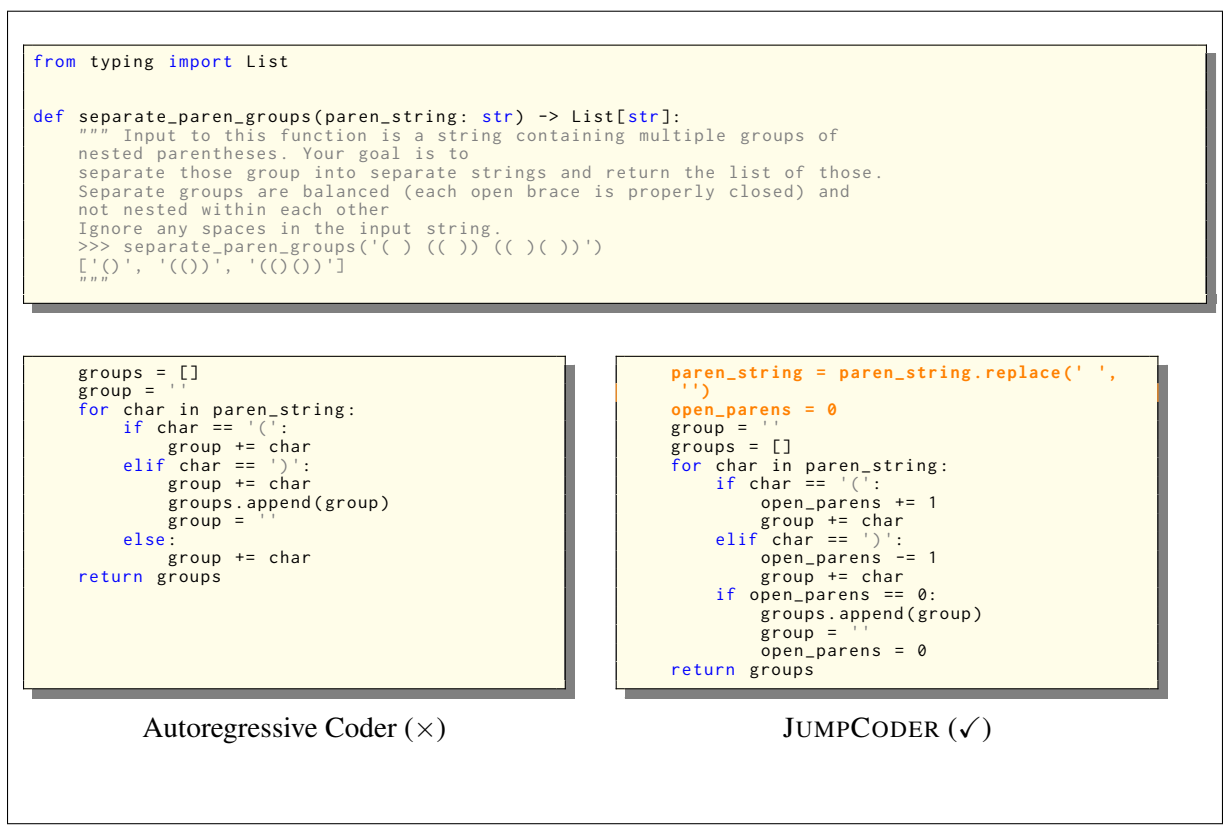

\begin{framed}

\begin{lstlisting}
from typing import List


def separate_paren_groups(paren_string: str) -> List[str]:
    """ Input to this function is a string containing multiple groups of 
    nested parentheses. Your goal is to
    separate those group into separate strings and return the list of those.
    Separate groups are balanced (each open brace is properly closed) and 
    not nested within each other
    Ignore any spaces in the input string.
    >>> separate_paren_groups('( ) (( )) (( )( ))')
    ['()', '(())', '(()())']
    """
\end{lstlisting}

\lstdefinestyle{autoregressioncoderstyle}{
  basicstyle=\fontsize{7}{7}\ttfamily,
  keywordstyle=\color{blue}, 
  morekeywords={if,else,while}, 
  framexbottommargin=35pt, 
  moredelim=[is][\color{red}\bfseries]{@}{@} 
}
\lstdefinestyle{jumpcoderstyle}{
  basicstyle=\fontsize{7}{7}\ttfamily,
  keywordstyle=\color{blue}, 
  morekeywords={if,else,while}, 
  moredelim=[is][\color{green!70!black}\bfseries]{@}{@},
  moredelim=[is][\color{orange}\bfseries]{*}{*}, 
}
\begin{minipage}[t][6cm]{0.45\columnwidth}
\begin{lstlisting}[style=autoregressioncoderstyle]
    groups = []
    group = ''
    for char in paren_string:
        if char == '(':
            group += char
        elif char == ')':
            group += char
            groups.append(group)
            group = ''
        else:
            group += char
    return groups
\end{lstlisting}
\centerline{Autoregressive Coder ($\times$)}
\end{minipage}
\hspace{20pt}
\begin{minipage}[t][6cm]{0.45\columnwidth}
\begin{lstlisting}[style=jumpcoderstyle]
*    paren_string = paren_string.replace(' ', '')*
*    open_parens = 0 *
    group = ''
    groups = []
    for char in paren_string:
        if char == '(':
            open_parens += 1
            group += char
        elif char == ')':
            open_parens -= 1
            group += char
        if open_parens == 0:
            groups.append(group)
            group = ''
            open_parens = 0
    return groups
\end{lstlisting}
\centerline{\model~(\checkmark)}
\end{minipage}

\end{framed}
\caption{An example program: In this example, the presence of spaces in the string requires preprocessing. Our approach successfully empowers the model to make modifications in filling the preprocessing code and introducing a necessary variable (\texttt{open\_parens}) for matching parentheses.}
\label{fig:good_example2}
\end{figure*}

\begin{figure*}[!htb]
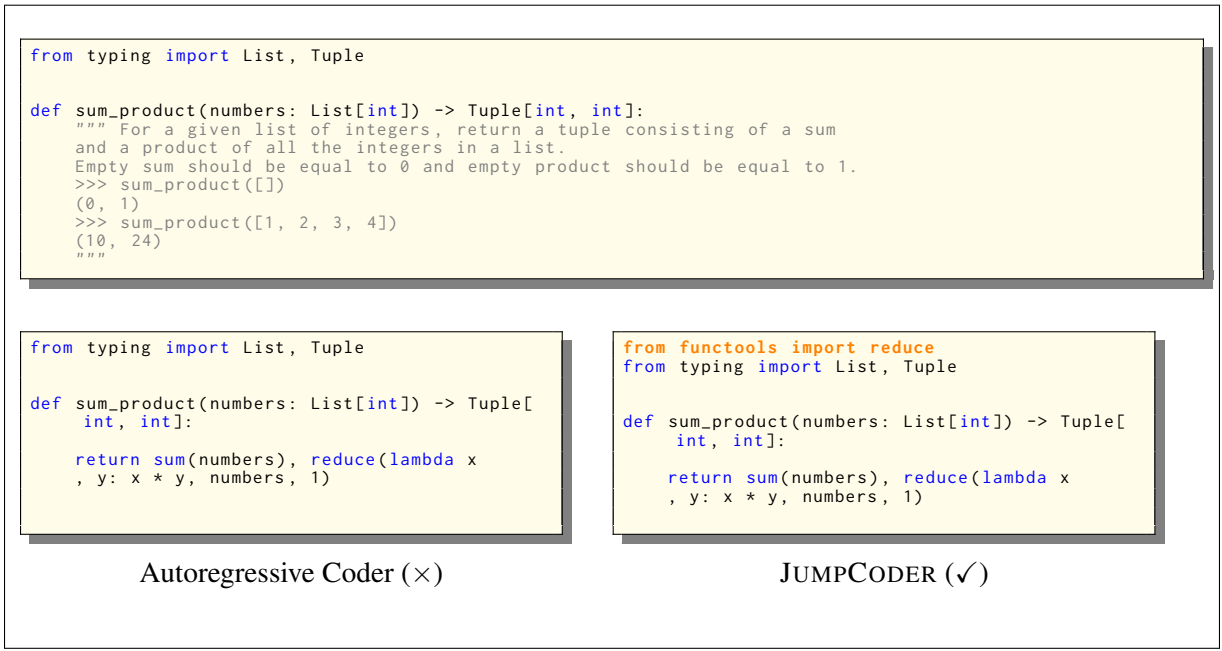

\begin{framed}

\begin{lstlisting}
from typing import List, Tuple


def sum_product(numbers: List[int]) -> Tuple[int, int]:
    """ For a given list of integers, return a tuple consisting of a sum 
    and a product of all the integers in a list.
    Empty sum should be equal to 0 and empty product should be equal to 1.
    >>> sum_product([])
    (0, 1)
    >>> sum_product([1, 2, 3, 4])
    (10, 24)
    """
\end{lstlisting}

\lstdefinestyle{autoregressioncoderstyle}{
  basicstyle=\fontsize{7}{7}\ttfamily,
  keywordstyle=\color{blue}, 
  morekeywords={if,else,while}, 
  framexbottommargin=14pt, 
  moredelim=[is][\color{red}\bfseries]{@}{@} 
}
\lstdefinestyle{jumpcoderstyle}{
  basicstyle=\fontsize{7}{7}\ttfamily,
  keywordstyle=\color{blue}, 
  morekeywords={if,else,while}, 
  framexbottommargin=7pt, 
  moredelim=[is][\color{green!70!black}\bfseries]{@}{@}, 
  moredelim=[is][\color{orange}\bfseries]{*^}{^*}, 
}
\begin{minipage}[t][4cm]{0.45\columnwidth}
\begin{lstlisting}[style=autoregressioncoderstyle]
from typing import List, Tuple


def sum_product(numbers: List[int]) -> Tuple[int, int]:

    return sum(numbers), reduce(lambda x
    , y: x * y, numbers, 1)
\end{lstlisting}
\centerline{Autoregressive Coder ($\times$)}
\end{minipage}
\hspace{20pt}
\begin{minipage}[t][4cm]{0.45\columnwidth}
\begin{lstlisting}[style=jumpcoderstyle]
*^from functools import reduce^*
from typing import List, Tuple


def sum_product(numbers: List[int]) -> Tuple[int, int]:

    return sum(numbers), reduce(lambda x
    , y: x * y, numbers, 1)
\end{lstlisting}
\centerline{\model~(\checkmark)}
\end{minipage}

\end{framed}
\caption{An example program: In this example, the code generated by the original model utilizes the \texttt{reduce} function from the \texttt{functools} library without importing the relevant function. Our approach ensures the model successfully imports the \texttt{reduce} function.}
\label{fig:good_example3}
\end{figure*}

\begin{figure*}[!htb]
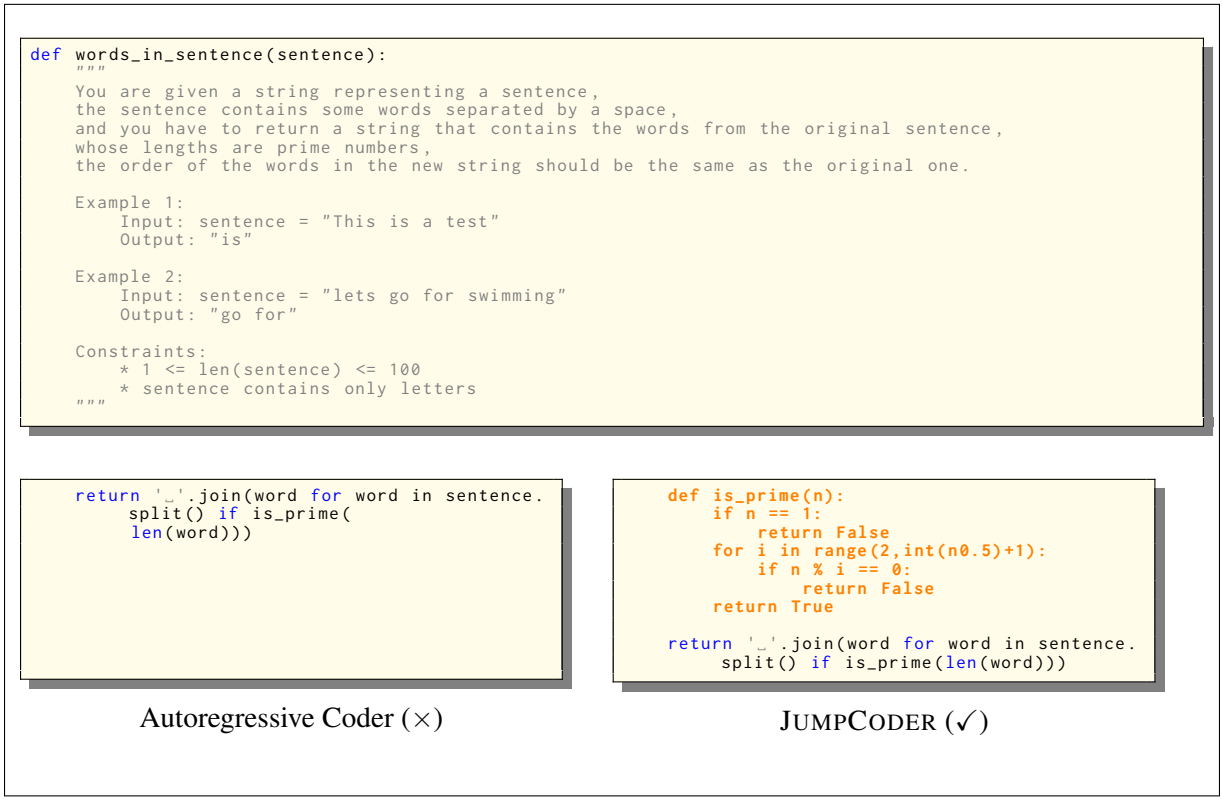

\begin{framed}

\begin{lstlisting}
def words_in_sentence(sentence):
    """
    You are given a string representing a sentence,
    the sentence contains some words separated by a space,
    and you have to return a string that contains the words from the original sentence,
    whose lengths are prime numbers,
    the order of the words in the new string should be the same as the original one.

    Example 1:
        Input: sentence = "This is a test"
        Output: "is"

    Example 2:
        Input: sentence = "lets go for swimming"
        Output: "go for"

    Constraints:
        * 1 <= len(sentence) <= 100
        * sentence contains only letters
    """
\end{lstlisting}

\lstdefinestyle{autoregressioncoderstyle}{
  basicstyle=\fontsize{7}{7}\ttfamily,
  keywordstyle=\color{blue}, 
  morekeywords={if,else,while}, 
  framexbottommargin=48pt, 
  moredelim=[is][\color{red}\bfseries]{@}{@} 
}
\lstdefinestyle{jumpcoderstyle}{
  basicstyle=\fontsize{7}{7}\ttfamily,
  keywordstyle=\color{blue}, 
  morekeywords={if,else,while}, 
  moredelim=[is][\color{green!70!black}\bfseries]{@}{@},
  moredelim=[is][\color{orange}\bfseries]{*}{*}, 
}
\begin{minipage}[t][4cm]{0.45\columnwidth}
\begin{lstlisting}[style=autoregressioncoderstyle]
    return ' '.join(word for word in sentence.split() if is_prime(
         len(word)))
\end{lstlisting}
\centerline{Autoregressive Coder ($\times$)}
\end{minipage}
\hspace{20pt}
\begin{minipage}[t][4cm]{0.45\columnwidth}
\begin{lstlisting}[style=jumpcoderstyle]
*    def is_prime(n):
        if n == 1:
            return False
        for i in range(2,int(n**0.5)+1):
            if n % i == 0:
                return False
        return True*
        
    return ' '.join(word for word in sentence.split() if is_prime(len(word)))
\end{lstlisting}
\centerline{\model~(\checkmark)}
\end{minipage}

\end{framed}
\caption{An example program: In this example, the code generated by the original model uses the undefined function \texttt{is\_prime(n)}. Our method enables the model to retroactively supplement the \texttt{is\_prime(n)} function.}
\label{fig:good_example4}
\end{figure*}

In this section, we first present several infills that were accepted (or rejected) by \model~during the Judging phase in \cref{tab:infill_example_good,tab:infill_example_bad}, to more intuitively demonstrate the workings of Judging. Then, we compare \model~with the traditional Autoregressive Coder through a few examples. Specifically, 
\cref{fig:good_example1,fig:good_example2,fig:good_example3,fig:good_example4} show some Python samples that are solved effectively by our framework, and \cref{fig:java_example1,fig:cpp_example1,fig:csharpe_example1} show some Java, C++ and C\# samples by our framework. Furthermore, \cref{fig:bad_example1} shows a sample that can be correctly solved by the autoregressive coder but is
degenerated by our framework. We mark the infills in \textbf{\textcolor{orange}{bold orange}}.

\lstdefinestyle{infill}{
  basicstyle=\fontsize{8}{8}\ttfamily,
  keywordstyle=\color{blue}, 
  morekeywords={if,else,while}, 
  moredelim=[is][\color{red}\bfseries]{@}{@}, 
  moredelim=[is][\color{orange}\bfseries]{^}{^},
  escapeinside={(*@}{@*)}, 
}

\begin{table*}[ht]
\centering
\small
\caption{Good infills \textit{accepted} by \model~during the Judging phase, along with the corresponding scores $V$. They can be broadly categorized into four types: Function, Reference, Variable, and Calculation.}
\label{tab:infill_example_good}
\begin{tabular}{ccc}
\toprule
Type & Example (infills are marked in \textbf{\textcolor{orange}{bold orange}}) & Infill score $V$ \\
\midrule
Function & \begin{minipage}[c]{10cm}\begin{lstlisting}[style=infill]
^def is_prime(n): ^
^   if n == 1:^
^       return False ^
^   ...^
return ' '.join(word for word in sentence.split() if is_prime(len(word)))
\end{lstlisting}
\end{minipage} & $+\infty$ \\
\midrule
Reference & \begin{minipage}[c]{10cm}\begin{lstlisting}[style=infill]
^from functools import reduce^
return sum(numbers), reduce(lambda x, y: x * y, numbers, 1)
\end{lstlisting}
\end{minipage} & $+\infty$ \\
\midrule
\multirow{5}[2]{*}{Variable} & \begin{minipage}[c]{10cm}\begin{lstlisting}[style=infill]
^curr_sum = 0^
for j in range(i, len(nums)):
\end{lstlisting}
\end{minipage} & 0.75 \\
\cmidrule{2-3}
& \begin{minipage}[c]{10cm}\begin{lstlisting}[style=infill]
^unique_chars = {}^
max_length = 0
\end{lstlisting}
\end{minipage} & 1.23 \\
\midrule
\multirow{5}[2]{*}{Calculation} & \begin{minipage}[c]{10cm}\begin{lstlisting}[style=infill]
^lst = [x for x in lst if len(x) % 2 == 0]^
lst = sorted(lst, key=lambda x: (len(x), x))
\end{lstlisting}
\end{minipage} & 1.38 \\
\cmidrule{2-3}
 & \begin{minipage}[c]{10cm}\begin{lstlisting}[style=infill]
^s = s.strip()^ # second infill
^s = s.replace(",", " ")^ # first infill
return s.split()
\end{lstlisting}
\end{minipage} & \begin{minipage}[c]{3cm}
\centering
first infill: 2.63 \\
second infill: 1.19
\end{minipage}\\
\bottomrule
\end{tabular}
\end{table*}

\begin{table*}[ht]
\centering
\small
\caption{Bad infills \textit{rejected} by \model~during the Judging phase, with the corresponding scores $V$. The first two examples violate syntax or coding standards, the third introduces incorrect calculations, and the fourth breaches semantics (\texttt{count += 1} should follow the \texttt{if} statement).}
\label{tab:infill_example_bad}
\begin{tabular}{cc}
\toprule
Example (infills are marked in \textbf{\textcolor{orange}{bold orange}}) & Infill score $V$ \\
\midrule
\begin{minipage}[c]{11cm}\begin{lstlisting}[style=infill]
if not test:
    return {}
^test = test.split()^
else:
\end{lstlisting}
\end{minipage} & -6.15 \\
\midrule
\begin{minipage}[c]{11cm}\begin{lstlisting}[style=infill]
if number + remaining >= need:
    return [number + remaining - need, need]
else:
    ^return [number + remaining, need - (number + remaining)]^
    return [number + remaining, 0]
\end{lstlisting}
\end{minipage} & -2.94 \\
\midrule
\begin{minipage}[c]{11cm}\begin{lstlisting}[style=infill]
if not arr:
    return True
^arr.sort()^
if len(arr) == 1:
\end{lstlisting}
\end{minipage} & -0.65 \\
\midrule
\begin{minipage}[c]{11cm}\begin{lstlisting}[style=infill]
count = 0
for sentence in S.split('.'):
    ^count += 1^
    if sentence.startswith('I'):
\end{lstlisting}
\end{minipage} & 0.19 \\
\bottomrule
\end{tabular}
\end{table*}

\clearpage


\begin{figure*}[!htb]
\lstdefinestyle{javastyle}{
    language=Java,
}
\begin{framed}

\begin{lstlisting}[style=javastyle]
import java.util.*;
import java.lang.reflect.*;
import org.javatuples.*;
import java.security.*;
import java.math.*;
import java.io.*;
import java.util.stream.*;
class Problem {
    // Input to this function is a string representing musical notes in a special ASCII format.
    // Your task is to parse this string and return array list of integers corresponding to how many beats does each not last.
    // Here is a legend:
    // 'o' - whole note, lasts four beats
    // 'o|' - half note, lasts two beats
    // '.|' - quater note, lasts one beat
    // >>> parseMusic(("o o| .| o| o| .| .| .| .| o o"))
    // (new ArrayList<Long>(Arrays.asList((long)4l, (long)2l, (long)1l, (long)2l, (long)2l, (long)1l, (long)1l, (long)1l, (long)1l, (long)4l, (long)4l)))
    public static ArrayList<Long> parseMusic(String music_string) {
\end{lstlisting}

\lstdefinestyle{autoregressioncoderstyle}{
  language = Java,
  basicstyle=\fontsize{7}{7}\ttfamily,
  keywordstyle=\color{blue}, 
  morekeywords={if,else,while}, 
  framexbottommargin=7pt, 
  moredelim=[is][\color{red}\bfseries]{@}{@} 
}
\lstdefinestyle{jumpcoderstyle}{
  language = Java,
  basicstyle=\fontsize{7}{7}\ttfamily,
  keywordstyle=\color{blue}, 
  morekeywords={if,else,while}, 
  framexbottommargin=28pt, 
  moredelim=[is][\color{green!70!black}\bfseries]{@}{@}, 
  moredelim=[is][\color{orange}\bfseries]{*}{*}, 
}
\begin{minipage}[t][6cm]{0.48\columnwidth}
\begin{lstlisting}[style=autoregressioncoderstyle]
ArrayList<Long> result = new ArrayList<Long>();
for (int i = 0; i < music_string.length(); i++) {
    if (music_string.charAt(i) == 'o') {
        result.add((long)4);
    } else if (music_string.charAt(i) == 'o' && music_string.charAt(i + 1) == '|') {
        result.add((long)2);
        i++;
    } else if (music_string.charAt(i) == '.' && music_string.charAt(i + 1) == '|') {
        result.add((long)1);
        i++;
    }
}
return result;
}
\end{lstlisting}
\centerline{Autoregressive Coder ($\times$)}
\end{minipage}
\hspace{20pt}
\begin{minipage}[t][6cm]{0.46\columnwidth}
\begin{lstlisting}[style=jumpcoderstyle]
*String[] notes = music_string.split(" ");*
ArrayList<Long> result = new ArrayList<Long>();
for (String note : notes) {
    if (note.equals("o")) {
        result.add((long)4);
    } else if (note.equals("o|")) {
        result.add((long)2);
    } else if (note.equals(".|")) {
        result.add((long)1);
    }
}
return result;
}
\end{lstlisting}
\centerline{\model~(\checkmark)}
\end{minipage}

\end{framed}
\caption{A Java example program: In this example, the presence of spaces in the string requires preprocessing. Our method successfully fills in the preprocessing code to split the input string into a string array.}
\label{fig:java_example1}
\end{figure*}

\begin{figure*}[!htb]
\lstdefinestyle{csharpestyle}{
    language=C++,
}
\begin{framed}

\begin{lstlisting}[style=csharpestyle]
using System;
using System.Numerics;
using System.Diagnostics;
using System.Collections.Generic;
using System.Linq;
using System.Text;
using System.Security.Cryptography;
class Problem {
    // Return a string containing space-delimited numbers starting from 0 upto n inclusive.
    // >>> StringSequence((0L))
    // ("0")
    // >>> StringSequence((5L))
    // ("0 1 2 3 4 5")
    public static string StringSequence(long n) {
\end{lstlisting}

\lstdefinestyle{autoregressioncoderstyle}{
  language = C++,
  basicstyle=\fontsize{7}{7}\ttfamily,
  keywordstyle=\color{blue}, 
  morekeywords={if,else,while}, 
  framexbottommargin=14pt, 
  moredelim=[is][\color{red}\bfseries]{@}{@} 
}
\lstdefinestyle{jumpcoderstyle}{
  language = C++,
  basicstyle=\fontsize{7}{7}\ttfamily,
  keywordstyle=\color{blue}, 
  morekeywords={if,else,while}, 
  framexbottommargin=7pt, 
  moredelim=[is][\color{green!70!black}\bfseries]{@}{@}, 
  moredelim=[is][\color{orange}\bfseries]{*}{*}, 
}
\begin{minipage}[t][4cm]{0.45\columnwidth}
\begin{lstlisting}[style=autoregressioncoderstyle]
        var sb = new StringBuilder();
        for (long i = 0; i <= n; i++) {
            sb.Append(i);
            sb.Append(" ");
        }
        return sb.ToString();
}
\end{lstlisting}
\centerline{Autoregressive Coder ($\times$)}
\end{minipage}
\hspace{20pt}
\begin{minipage}[t][4cm]{0.45\columnwidth}
\begin{lstlisting}[style=jumpcoderstyle]
        var sb = new StringBuilder();
        for (long i = 0; i <= n; i++) {
            sb.Append(i);
            sb.Append(" ");
        }
        *sb.Length--;*
        return sb.ToString();
}
\end{lstlisting}
\centerline{\model~(\checkmark)}
\end{minipage}

\end{framed}
\caption{A C\# example program: In this example, our approach incorporates an additional step to remove the space at the end of the string.}
\label{fig:csharpe_example1}
\end{figure*}

\begin{figure*}[!htb]
\lstdefinestyle{cppstyle}{
    language=C++,
}
\begin{framed}

\begin{lstlisting}[style=cppstyle]
#include<assert.h>
#include<bits/stdc++.h>
// You are given a string s.
// if s[i] is a letter, reverse its case from lower to upper or vise versa, 
// otherwise keep it as it is.
// If the string contains no letters, reverse the string.
// The function should return the resulted string.
// Examples
// >>> solve(("1234"))
// ("4321")
// >>> solve(("ab"))
// ("AB")
// >>> solve(("#a@C"))
// ("#A@c")
std::string solve(std::string s) {
\end{lstlisting}

\lstdefinestyle{autoregressioncoderstyle}{
  language = C++,
  basicstyle=\fontsize{7}{7}\ttfamily,
  keywordstyle=\color{blue}, 
  morekeywords={if,else,while}, 
  framexbottommargin=56pt, 
  moredelim=[is][\color{red}\bfseries]{@}{@} 
}
\lstdefinestyle{jumpcoderstyle}{
  language = C++,
  basicstyle=\fontsize{7}{7}\ttfamily,
  keywordstyle=\color{blue}, 
  morekeywords={if,else,while}, 
  framexbottommargin=7pt, 
  moredelim=[is][\color{green!70!black}\bfseries]{@}{@}, 
  moredelim=[is][\color{orange}\bfseries]{*}{*}, 
}
\begin{minipage}[t][6cm]{0.45\columnwidth}
\begin{lstlisting}[style=autoregressioncoderstyle]
    std::string result;
    for (char c : s) {
        if (c >= 'a' && c <= 'z') {
            result += c - 'a' + 'A';
        } else if (c >= 'A' && c <= 'Z') {
            result += c - 'A' + 'a';
        } else {
            result += c;
        }
    }
    return result;
}
\end{lstlisting}
\centerline{Autoregressive Coder ($\times$)}
\end{minipage}
\hspace{20pt}
\begin{minipage}[t][6cm]{0.45\columnwidth}
\begin{lstlisting}[style=jumpcoderstyle]
    *bool has_letter = false;*
    std::string result;
    for (char c : s) {
        if (c >= 'a' && c <= 'z') {
            result += c - 'a' + 'A';
            has_letter = true;
        } else if (c >= 'A' && c <= 'Z') {
            result += c - 'A' + 'a';
            has_letter = true;
        } else {
            result += c;
        }
    }
    if (!has_letter) {
        std::reverse(result.begin(), result.end());
    }
    return result;
}
\end{lstlisting}
\centerline{\model~(\checkmark)}
\end{minipage}

\end{framed}
\caption{A C++ example program: In this example, our approach retrospectively defines a necessary variable (\texttt{has\_letter}) which records the presence of a letter.}
\label{fig:cpp_example1}
\end{figure*}

\begin{figure*}[!htb]
\begin{framed}

\begin{lstlisting}
def digitSum(s):
    """Task
    Write a function that takes a string as input and returns the sum of the upper characters only'
    ASCII codes.

    Examples:
        digitSum("") => 0
        digitSum("abAB") => 131
        digitSum("abcCd") => 67
        digitSum("helloE") => 69
        digitSum("woArBld") => 131
        digitSum("aAaaaXa") => 153
    """
\end{lstlisting}

\lstdefinestyle{autoregressioncoderstyle}{
  basicstyle=\fontsize{7}{7}\ttfamily,
  keywordstyle=\color{blue}, 
  morekeywords={if,else,while}, 
  framexbottommargin=56pt, 
  moredelim=[is][\color{red}\bfseries]{@}{@} 
}
\lstdefinestyle{jumpcoderstyle}{
  basicstyle=\fontsize{7}{7}\ttfamily,
  keywordstyle=\color{blue}, 
  morekeywords={if,else,while}, 
  framexbottommargin=49pt, 
  moredelim=[is][\color{red}\bfseries]{@}{@} ,
    moredelim=[is][\color{orange}\bfseries]{*}{*}, 
}
\begin{minipage}[t][4cm]{0.45\columnwidth}
\begin{lstlisting}[style=autoregressioncoderstyle]
    sum = 0
    for c in s:
        if c.isupper():
            sum += ord(c)
    return sum
\end{lstlisting}
\centerline{Autoregressive Coder (\checkmark)}
\end{minipage}
\hspace{20pt}
\begin{minipage}[t][4cm]{0.45\columnwidth}
\begin{lstlisting}[style=jumpcoderstyle]
   * s = s.upper()*
    sum = 0
    for c in s:
        if c.isalpha():
            sum += ord(c)
    return sum
\end{lstlisting}
\centerline{\model~($\times$)}
\end{minipage}

\end{framed}
\caption{An example program: In this example, we illustrate that the infilling model may exhibit a certain level of misunderstanding, filling in possibly erroneous code.}
\label{fig:bad_example1}
\end{figure*}

\end{document}